\pdfoutput=1

\documentclass[11pt]{article}

\usepackage[final]{acl}

\usepackage{times}
\usepackage{latexsym}
\usepackage[T1]{fontenc}
\usepackage[utf8]{inputenc}
\usepackage{microtype}
\usepackage{inconsolata}
\usepackage{tcolorbox}
\usepackage{siunitx, threeparttable, booktabs, multirow, xcolor, colortbl, graphicx}
\newcommand{\besttwo}[1]{\cellcolor{yellow!25}#1}
\newcommand{\bestone}[1]{\cellcolor{orange!20}#1}
\newcommand{\bestthree}[1]{\cellcolor{cyan!20}#1}

\usepackage{tabularx}  

\newcolumntype{L}{>{\raggedright\arraybackslash}X}

\usepackage{graphicx}
\usepackage{paralist,xcolor}
\usepackage{bm}
\usepackage{multirow}
\usepackage[export]{adjustbox}

\usepackage{colortbl}
\definecolor{lightgray}{gray}{.9}
\definecolor{deepgray}{gray}{.8}

\usepackage{threeparttable}
\newcolumntype{I}{!{\vrule width 1pt}}
\makeatletter
\newcommand{\thickhline}{%
    \noalign {\ifnum 0=`}\fi \hrule height 1pt
    \futurelet \reserved@a \@xhline
}
\makeatother
\definecolor{mygray}{gray}{.9}
\usepackage{float}
\usepackage[ruled, vlined]{algorithm2e}
\usepackage{pifont}
\usepackage{ragged2e}
\usepackage{paralist}
\usepackage[toc,page]{appendix}
\usepackage{url}
\usepackage{subcaption}
\newcolumntype{g}{>{\columncolor{gray!20}}c} 

\usepackage{lipsum} 
\usepackage{subcaption} 

\usepackage{array, adjustbox}
\usepackage{booktabs}
\usepackage{amssymb} 
\usepackage{paralist}
\usepackage{booktabs,ltablex,adjustbox,array}
\renewcommand{\arraystretch}{1.1}
\usepackage{siunitx}
\usepackage{booktabs, multirow, threeparttable}
\usepackage{siunitx}
\usepackage{rotating}   
\usepackage{booktabs}
\usepackage{amsmath}
\usepackage{mathtools}
\usepackage{amsthm}
\usepackage{amssymb}
\usepackage{bm}
\usepackage{multirow}
\usepackage{graphicx}  
\usepackage{booktabs}  
\usepackage{subcaption}  
\usepackage[normalem]{ulem}
\usepackage{afterpage}
\usepackage{threeparttable}

\newcommand{\wei}[1]{\textcolor{orange}{\#Wei:~#1\#}}

\newcommand{\zw}[1]{\textcolor{blue}{\#Zewen:~#1\#}}

\title{SymbolBench: Can Large Language Models Adequately Perform Symbolic Reasoning Over Time Series?}

\author{
Zewen Liu$^{1}$, Juntong Ni$^{1}$, Xianfeng Tang$^{2}$,
Max S.Y. Lau$^{1}$, Qi He$^{3}$,  Wenpeng Yin$^{4}$, Wei Jin$^{1}$,
\\
$^{1}$Emory University, $^{2}$Amazon, $^{3}$Microsoft, $^{4}$Penn State University \\
\texttt{\{zewen.liu, juntong.ni, msy.lau, wei.jin\}@emory.edu}, \\
\texttt{qhe@microsoft.com}; \texttt{xianfengtang@outlook.com}; \texttt{wenpeng@psu.edu}
}

\begin{document}
\maketitle
\begin{abstract}

Uncovering hidden symbolic laws from time series data, as an aspiration dating back to Kepler's discovery of planetary motion, remains a core challenge in scientific discovery and artificial intelligence. While Large Language Models show promise in structured reasoning tasks, their ability to infer interpretable, context-aligned symbolic structures from time series data is still underexplored. To systematically evaluate this capability, we introduce \textbf{SymbolBench}, a comprehensive benchmark designed to assess symbolic reasoning over real-world time series across three tasks: \textit{multivariate symbolic regression, Boolean network inference}, and \textit{causal discovery}. Unlike prior efforts limited to simple algebraic equations, SymbolBench spans a diverse set of symbolic forms with varying complexity. We further propose a unified framework that integrates LLMs with genetic programming to form a closed-loop symbolic reasoning system. Our empirical results reveal key strengths and limitations of current models, highlighting the importance of combining domain knowledge, context alignment, and reasoning structure to improve LLMs in automated scientific discovery. \url{https://github.com/nuuuh/SymbolBench}

\end{abstract}

\section{Introduction}
Centuries ago, Johannes Kepler revolutionized our understanding of the cosmos by discovering the laws of planetary motion~\cite{gentner2002analogy}. Through meticulous analysis and rigorous reasoning over astronomical observations, captured as time series data of planetary positions, Kepler derived precise mathematical relationships that explained the complex, time-dependent dynamics of celestial orbits. Automating such a challenging process, where \textbf{hidden symbolic laws are extracted from time series}, is a long-standing aspiration in artificial intelligence~\cite{reddy2025towards}. 
However, achieving this automation presents fundamental reasoning-driven challenges, as time series data encapsulates dynamic behaviors and temporal dependencies that demand abstraction, generalization, and reasoning beyond mere pattern recognition to uncover underlying symbolic structures.

\begin{figure}[t]
    \centering
    \includegraphics[width=\linewidth]{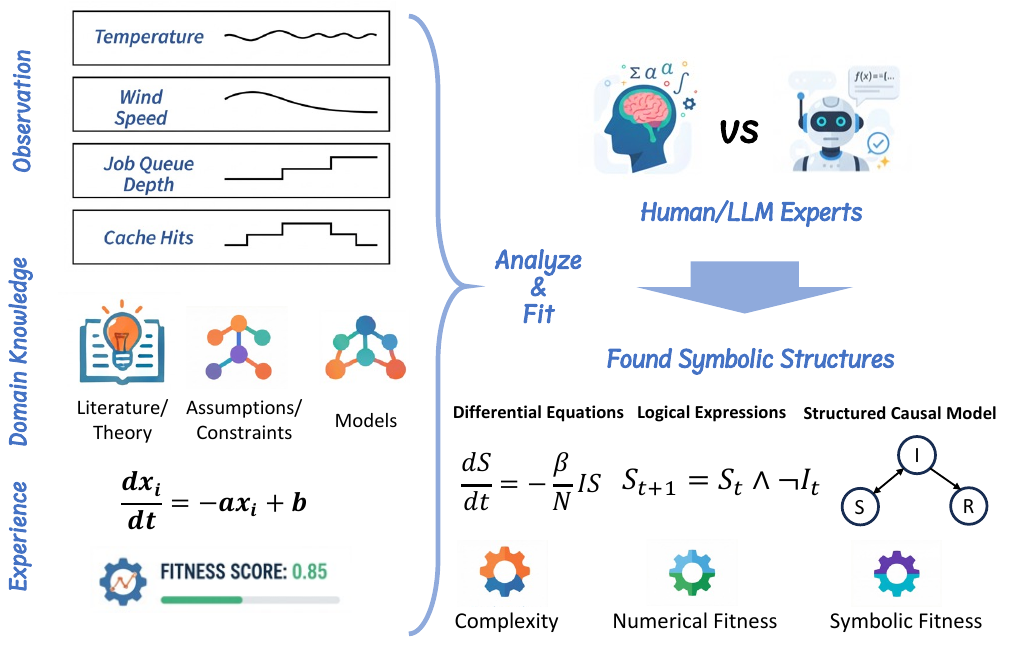}
    \vskip -1em
    \caption{Symbolic structure discovery from time series.}
    \label{fig: demo}
    \vskip -1.5em
\end{figure}

Recent advances in Large Language Models (LLMs) and Multimodal LLMs (MLLMs) show strong performance in complex reasoning tasks~\cite{wang2024exploring}, but their ability to extract symbolic laws from time series remains underexplored. Traditional methods in symbolic regression, such as genetic programming~\cite{makke2024interpretable}, often prioritize data fit at the expense of interpretability. Meanwhile, recent attempts~\cite{merler2024context, li2024mllm, shojaee2024llm} to use LLMs in this domain have been limited; they employ them merely as 
domain-agnostic function generators, overlooking the models' core potential for deeper, theory-aligned reasoning. This shallow integration often results in proposed equations that lack contextual relevance. Additionally, most work focuses on algebraic expressions, neglecting other symbolic forms like logical formulas~\cite{zhang2024logicgep} and causal relations~\cite{assaad2022survey}.




To address these gaps and provide further insights into the symbolic reasoning ability of LLMs for time series, a comprehensive benchmark is urgently required. Our work is guided by four key objectives: (a) \textbf{Real-world relevance:} To use real-world time series with ground-truth symbolic structures. (b) \textbf{Task difficulty:} To incorporate a diverse form of symbolic structures with varying complexity. (c) \textbf{Scale and balance:} To ensure a sufficiently large and balanced sample distribution across tasks.
(d) \textbf{Unified framework:} To provide a unified framework to execute various tasks and establish connections with task-specific baselines.

To realize these objectives, we present three primary contributions. 
\textit{\textbf{(I) SymbolBench}}, a comprehensive benchmark designed to rigorously evaluate the symbolic reasoning capabilities of LLMs over time series with rich real-world contextual descriptions. It uniquely spans three core tasks covering major types of time series data: (a)~\textbf{Multivariate symbolic regression} for continuous data to recover complex equations like coupled Ordinary Differential Equations (ODEs). (b)~\textbf{Boolean network inference}~\cite{zhang2024logicgep} for discrete systems to identify logical rules. (c)~\textbf{Causal discovery}~\cite{assaad2022survey} for multivariate data to uncover structured causal graphs. Each task includes challenging, real-world examples from domains such as biology, physics, and healthcare, with varying dimensionality and difficulty. To ensure both quality and coverage, we curate representative subsets from large databases, yielding a benchmark that is broader, more challenging, and more balanced than prior efforts.
\textit{\textbf{(II) Unified Symbolic Reasoning Framework}} that enables the context-aware and knowledge-rich LLMs to play the dual role of predictors and judges in the process of hypothesis generation, testing, and refinement, while optionally including efficient genetic programming tools in a hybrid way.
\textit{\textbf{(III) Critical empirical insights}} into the strengths and limitations of LLMs and MLLMs in temporal symbolic reasoning: (a) LLMs outperform traditional baselines on multivariate symbolic regression and causal discovery but lag in Boolean network inference; (b) they demonstrate task-level reasoning, with increased test-time compute yielding moderate gains; (c) contextual information enhances both performance and the discovery of generalizable symbolic structures; (d) integrating LLMs with genetic programming in complementary roles further improves results.

\section{Related Work}
\noindent\textbf{Symbolic Expression Discovery.}
Symbolic regression (SR) aims to recover interpretable equations from time series data. Classical methods like Genetic Programming and sparse optimization (e.g., SINDy~\cite{brunton2016discovering}, PySR~\cite{cranmer2024pysr}) prioritize accuracy and parsimony but face scalability challenges. Recent deep learning models (e.g., ODEformer~\cite{d2023odeformer}, TPSR~\cite{shojaee2023transformer}) improve efficiency by treating SR as a translation task, though they require pretraining and lack iterative refinement. Related fields like Boolean network inference and causal discovery also seek symbolic structures from time series. More details are in Appendix~\ref{appendix: related_work}.


\noindent\textbf{LLM Symbolic Reasoning.}
LLMs demonstrate strong in-context learning and logical reasoning abilities~\cite{ahn2024large, wang2023review}, and have been applied to time series analysis~\cite{fang2024large}. However, most prior work, such as ChatTS~\cite{xie2024chatts}, LLM-ABB~\cite{carson2024llm}, and TemporalGPT~\cite{zhang2025tempogpt}, focuses on the compression or reasoning at the \textbf{linguistic level}, summarizing trends, or producing descriptive insights, rather than uncovering hidden symbolic structures. By contrast, our tasks target recovery of scientific equations, logical rules, and causal graphs, which demand abstraction, contextual alignment, and explanatory power. 
Existing attempts in symbolic regression often reduce LLMs to function generators~\cite{merler2024context, li2024mllm, shojaee2024llm}, yielding high data fit but low context-alignment, and largely overlook symbolic forms beyond algebraic functions~\cite{zhang2024logicgep, assaad2022survey}, e.g., Boolean networks~\cite{zhang2024logicgep} and causal relations~\cite{assaad2022survey}.
\section{SymbolBench Dataset}


To rigorously assess symbolic reasoning over time series, we curate SymbolBench by systematically constructing datasets that combine real-world scientific systems with carefully controlled simulation and contextual annotation. Across all tasks, our construction follows a unified pipeline: (i) identify and extract well-formed ground-truth symbolic structures from established scientific repositories, (ii) generate time series trajectories through simulation under varied initial conditions, (iii) enrich each sample with available domain and variable-level metadata, and (iv) balance dataset distributions to avoid bias toward low-dimensional or trivial cases. Eventually, the goal of each task is to recover the symbolic structure from the corresponding time series. This design allows SymbolBench to address three key dataset objectives: real-world relevance, task difficulty, and scale and balance.


\noindent\textbf{(a) Coupled Differential Equations (CDEs)}.  
We extend ODEBench~\cite{d2023odeformer} by incorporating additional high-dimensional systems, addressing its imbalance of a few four-dimensional examples. Using the Physiome library~\cite{yu2011physiome}, which includes over 500 cell-dynamics models, we exclude systems with discrete states, external inputs, or excessively long coefficient lists, focusing on continuous dynamics with symbolic structural complexity. For each selected system, we integrate the governing equations under multiple initial conditions and parameter settings to generate diverse continuous trajectories. Each sample includes textual metadata (variable descriptions and domain labels, e.g., physics, biology)for contextual grounding. The final CDE dataset comprises over 150 balanced samples spanning one- to four-dimensional systems across diverse domains such as physical mechanics, electronics, social dynamics, epidemiology, calcium signaling, neurobiology, and cardiovascular circulation. Each sample uses 150 data points for fitting and evaluation.

\noindent\textbf{(b) Boolean Networks (BNs)}. To represent discrete dynamical systems, we extract Boolean networks from BioDivine~\cite{pastva2023repository}, a repository of biologically inspired gene regulatory and signaling networks. Each variable is represented as a binary node whose state evolves via logical update rules (See Table~\ref{tab: dataset_examples} in the appendix). We select networks with fewer than 20 variables and filter out those with trivial dynamics (e.g., all states collapsing to a single attractor in one step). We simulate temporal trajectories from given initial conditions, yielding sequences of binary state transitions. Independent variables are randomly assigned values to ensure variability. Alongside the logical rules, we provide domain descriptions and variable annotations, ensuring interpretability in the biological context. Our curated subset comprises 65 Boolean networks across diverse regulatory settings. For each sample, we provide 30 pairs of transitions by randomly varying the initial conditions.

\noindent\textbf{(c) Structured Causal Models (SCMs)}. Beyond algebraic and logical dynamics, we include causal structures by deriving SCMs using functional analysis, which extracts the variable dependencies of samples from two sources: (a) our curated CDEs dataset, and (b) additional multivariate dynamical systems from the Physiome database. Each SCM is formalized as a directed graph where edges encode causal dependencies with explicit time lags. To maintain interpretability, we annotate variables with their scientific roles (e.g., ion concentrations, membrane potentials) and domains. This results in 190 SCM samples covering complex physiological and physical systems. For each sample, we sample 150 data points based on the ground truth CDE.

\noindent\textbf{\underline{Stress Tests:}} 
Beyond standard in-distribution splits, SymbolBench includes evaluations under \textbf{out-of-distribution (OOD)} and \textbf{noisy} settings created by systematically varying initial conditions and control parameters outside the fitting range, and adding noise to the time series during fitting.
This design mirrors real-world challenges where scientific systems rarely operate under a single controlled regime or without random external noise, thereby testing a model’s ability to generalize to unseen but realistic scenarios rather than simply interpolating observed, noise-free patterns. More details of the dataset are provided in Appendix~\ref{appendix: dataset_examples}.

\begin{table}[t]
\centering
\caption{Comparison of SymbolBench with existing benchmarks for LLMs across symbolic structures, evaluation setups, contextual data types, and reasoning.}
\vskip -0.8em
\label{tab:symbolbench_comparison}

\scalebox{0.6}{
\begin{tabular}{lccc g}     
\toprule
\textbf{Feature} & \textbf{LLM-SRBench} & \textbf{ODEBench} & \textbf{RealTCD} & \textbf{Ours} \\
\midrule
\multicolumn{4}{l}{\textit{Symbolic Structures}} & \cellcolor{gray!20} \\[-0.3em]
\quad Scientific Eq. & 128 & 63 & -- & 156 \\
\quad Logical Exp.   & --  & -- & -- &  65 \\
\quad SCM            & --  & -- &  2 & 190 \\
\midrule
\multicolumn{4}{l}{\textit{Evaluation}} & \cellcolor{gray!20} \\[-0.3em]
\quad ID/OOD    & $\checkmark$ & $\checkmark$ & -- & $\checkmark$ \\
\quad Reasoning & -- & -- & -- & $\checkmark$ \\
\midrule
\multicolumn{4}{l}{\textit{Context}} & \cellcolor{gray!20} \\[-0.3em]
\quad Textual    & $\checkmark$ & -- & -- & $\checkmark$ \\
\quad Multimodal & -- & -- & -- & $\checkmark$ \\
\bottomrule
\end{tabular}}
\vskip -1em
\end{table}

\begin{figure*}[t]
    \centering
    \includegraphics[width=0.95\linewidth]{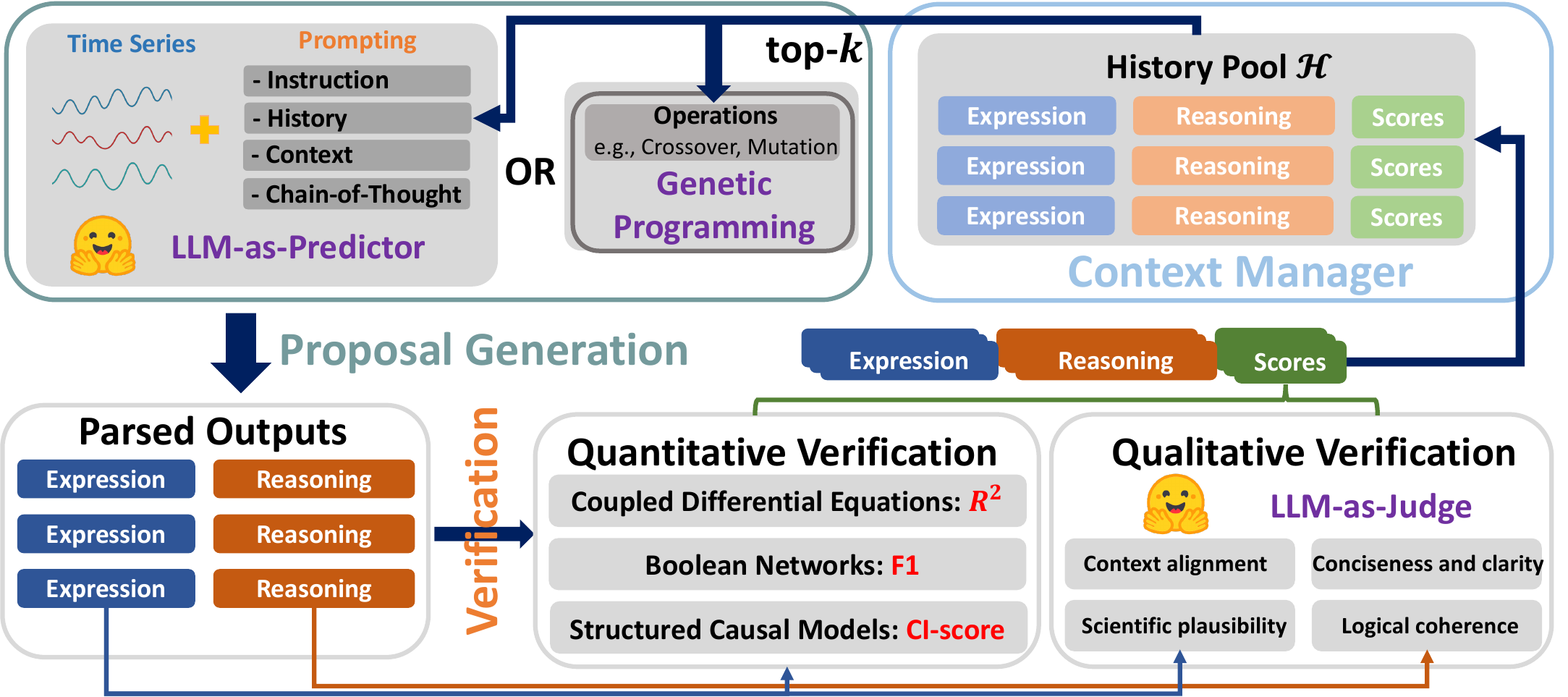}
    \caption{Iterative refinement framework with hybrid design. Candidate proposals are generated using either an LLM-as-Predictor or genetic programming operations. Each round of candidates undergoes quantitative and qualitative evaluation via validation tools and an LLM-as-Judge. Scored candidates are stored in a history pool, and a context manager decides contextual information for the next round.}
    \label{fig:framework}
    \vskip -1.5em
\end{figure*}

\section{SymbolBench Reasoning Framework}
Given an input time series with T time points \(\{\mathbf{x}^i\}_{i=1}^{T}\), where $\mathbf{x}^i \in \mathbb{R}^{D}$ and D is the number of dimensions, our framework aims to generate a subset of symbolic structures for each time series. Figure~\ref{fig:framework} sketches the closed-loop workflow that combines LLMs into the automatic pipeline. The process starts with \emph{Proposal Generation} (Sec.~\ref{subsec:PG}), which generates candidate expressions alongside their reasoning path if available. Then, the raw output is cleaned and used for \emph{Verification} (Sec.~\ref{subsec:verifier}), which assigns scores for each candidate. Candidates are then stored in a history pool. Finally, before the next round of generation, the \emph{Context Manager} (Sec.~\ref{subsec:context-manager}) extracts candidates from the history pool based on certain rules and provides them as the context for the next round proposal generation. The loop repeats until either the stopping criterion is met or the budget exceeds the limit.

\subsection{Proposal Generation}
\label{subsec:PG}
\noindent\textbf{LLM-as-Predictor.}
To leverage LLMs’ embedded knowledge, prior work has explored direct equation generation, which we term \textit{LLM-as-Predictor}. To evaluate both model capability and the effects of iterative refinement, context, and chain-of-thought reasoning, we design four prompting strategies for LLMs/MLLMs: 
\textbf{(a) Naive}, which generates expressions without contextual or historical input; 
\textbf{(b) Base}, which adds filtered prior expressions from the history pool; 
\textbf{(c) Context}, which further incorporates relevant context such as variable descriptions; 
\textbf{(d) CoT}, which extends Context with step-by-step reasoning, which is recorded for qualitative verification. 
Full prompt details are in Appendix~\ref{appendix: prompts}.

\vskip 0.2em
\noindent\textbf{Hybrid Method.}
While generating symbolic structures is central to the pipeline, this step can also leverage genetic programming operations. For CDEs and BNs, this entails applying crossover on expression trees to create new candidates. As LLMs can support other components, e.g., verification, we refer to this integration as the \textit{Hybrid Method} (Figure~\ref{fig: hybrid_method_}), as discussed in Appendix~\ref{appendix: hybrid_method}.

\begin{figure}[htb]
  \centering
  \subfloat[Genetic Programming + LLM-as-Judge.\label{fig: hybrid_gp_}]{
    \includegraphics[width=0.42\columnwidth]{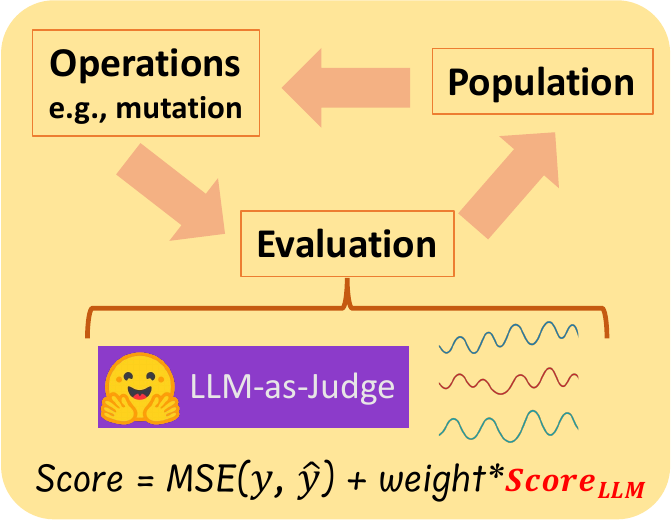}}
  \hfil
  \subfloat[Genetic Programming + LLM-as-Predictor.\label{fig: hybrid_llm_}]{
    \includegraphics[width=0.52\columnwidth]{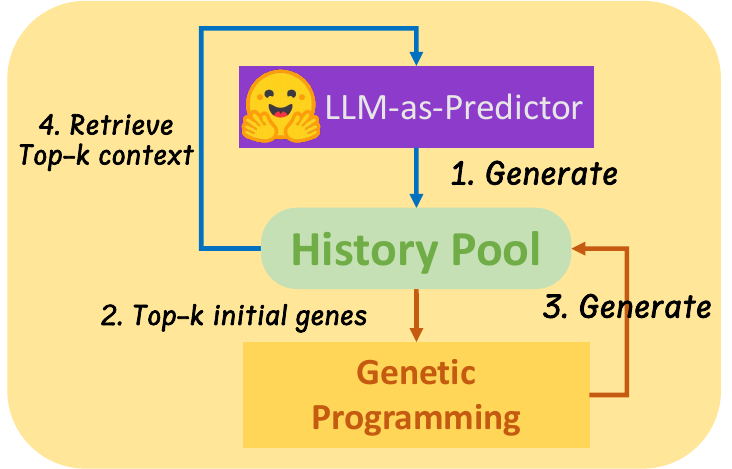}}
  \caption{Hybrid method.}
  \label{fig: hybrid_method_}
  \vskip -1em
\end{figure}

\subsection{Verification}
\label{subsec:verifier}
Verification in symbolic structure discovery assesses three key aspects: numerical fitness, symbolic fitness, and context alignment. However, unlike numerical fitness, symbolic fitness and context alignment are hard to quantify straightforwardly. To address this, we introduce a dual-verification strategy combining quantitative metrics with rubric-based qualitative evaluation with \textit{LLMs-as-Judges}.

\subsubsection{Quantitative Verification}
\noindent\textbf{(a)} For CDEs, we simulate time series from predicted functions and assess numerical fitness via the averaged coefficient of determination ($R^2$), with expression complexity measured by the number of operations. 
\noindent\textbf{(b)} For Boolean networks, we simulate transitions from predicted networks and evaluate predictive performance using the macro-averaged F1 score over $T$ transitions; expression complexity is computed as in CDEs. 
\noindent\textbf{(c)} For structural causal models, where the true generative process is unknown, we use the \emph{Conditional Independence score (CI-score)} as a proxy for structural fidelity, measuring dependence between each child and its parents conditioned on the remaining parents. Lower CI-scores indicate stronger conditional independence and higher structural plausibility.

\subsubsection{Qualitative Verification}
To evaluate the symbolic quality of generated expressions, we apply a rubric-based scoring from 1 (poor) to 5 (excellent). Each score is assigned by an LLM judge~\cite{hao2024llm} following a standardized rubric with four criteria.
\textbf{(a) Context alignment}: consistency with time series data and contextual descriptions; 
\textbf{(b) Scientific plausibility}: adherence to physical laws or domain constraints; 
\textbf{(c) Conciseness and clarity}: readability and succinctness of reasoning; 
\textbf{(d) Logical coherence}: stepwise consistency and sound derivation.

\subsection{Context Managing}
\label{subsec:context-manager}
The Context Manager supplies relevant information to the Proposal Generator during iterative refinement. Depending on the strategy, it selects an appropriate context and maintains a \emph{history pool} \(\mathcal{H}\) storing prior expressions, reasoning paths, and verification scores in a structured DataFrame, with duplicates removed. We adopt a simple ranking-based approach that selects the top-\(k\) highest-scoring candidates for further refinement.

\section{Experiment}

\begin{table*}[ht]
\centering
\caption{Symbolic regression results for CDEs across 4 dimensions. We use percentage for $SR^2$ and $ACC_{0.9}$. \textcolor{orange}{Orange} and \textcolor{yellow}{Yellow} mark the first and second place, respectively. 
Statistically significant improvements over the best baseline (paired t-test, $p$ < 0.05) are \underline{underlined}.}
\vskip -0.8em
\label{tab: CDEs}
\renewcommand{\arraystretch}{1}
\setlength{\tabcolsep}{0.5pt}
\begin{threeparttable}
\fontsize{7.5}{9}\selectfont
\begin{tabular}{
c l l
*{3}{S[table-format=3.2]}
*{2}{S[table-format=3.2]}
*{2}{S[table-format=3.2]}
*{2}{S[table-format=3.2]}
*{2}{S[table-format=3.2]}
*{2}{S[table-format=3.2]}
*{2}{S[table-format=3.2]}
}
\toprule
\multirow{2}{*}{Dim} & \multirow{2}{*}{} & \multirow{2}{*}{Metric}
  & \multicolumn{3}{c}{Baselines}
  & \multicolumn{2}{c}{Qwen2.5}
  & \multicolumn{2}{c}{Llama3.2}
  & \multicolumn{2}{c}{Mathstral}
  & \multicolumn{2}{c}{4o-text}
  & \multicolumn{2}{c}{4o-image}
  & \multicolumn{2}{c}{ChatTS} \\
\cmidrule(lr){4-6}\cmidrule(lr){7-8}\cmidrule(lr){9-10}\cmidrule(lr){11-12}
\cmidrule(lr){13-14}\cmidrule(lr){15-16}\cmidrule(lr){17-18}
 &  & 
 & {Pro.} & {PySR} & {ODE.}
 & {Ctx} & {CoT}
 & {Ctx} & {CoT}
 & {Ctx} & {CoT}
 & {Ctx} & {CoT}
 & {Ctx} & {CoT}
 & {Ctx} & {CoT} \\
\midrule
\multirow{6}{*}{\rotatebox[origin=c]{90}{Dim=1}}
  & \multicolumn{2}{l}{Complexity $\downarrow$}                   & 5.06 & 2.91 & 4.68 & 5.34 & 3.91 & \bestone{1.89} & \besttwo{1.97} & 2.63 & 2.69 & 2.31 & 2.36 & 2.37 & 2.09 & 3.29 & 3.00 \\
  & \multicolumn{2}{l}{Symbolic Prox. $\downarrow$}               & 4.86 & \bestone{3.54} & 5.18 & 5.57 & 4.89 & 4.03 & 4.14 & 3.86 & 4.09 & 3.89 & \besttwo{3.79} & 3.97 & 4.06 & 4.83 & 4.37 \\
  & \multirow{2}{*}{ID}  & $SR^2$                               & 95.50 & 96.90 & 83.40 & \bestone{\underline{99.00}} & \besttwo{\underline{97.40}} & 95.10 & 94.90 & 96.20 & 96.20 & 92.80 & 95.40 & 95.70 & 95.30 & 95.10 & 97.00 \\
  &                       & $ACC_{0.9}$                           & 91.40 & \besttwo{97.10} & 73.50 & \bestone{\underline{100.00}} & \besttwo{97.10} & 91.40 & 91.40 & \besttwo{97.10} & \besttwo{97.10} & 94.30 & 97.00 & \besttwo{97.10} & \besttwo{97.10} & 91.40 & \besttwo{97.10} \\
  & \multirow{2}{*}{OOD} & $SR^2$                               & \besttwo{74.10} & \bestone{88.20} & 54.10 & 61.80 & 61.80 & 69.90 & 64.60 & 71.00 & 66.90 & 70.80 & 66.20 & 71.50 & 62.90 & 63.80 & 70.30 \\
  &                       & $ACC_{0.9}$                           & \besttwo{65.70} & \bestone{85.70} & 47.10 & 52.90 & 51.40 & 51.40 & 48.60 & 62.90 & 60.00 & 54.30 & 51.50 & 54.30 & 45.70 & 51.40 & 54.30 \\
\cmidrule(lr){2-18}
 & \multicolumn{2}{l}{Avg. ID Rank $\downarrow$}  & 10.5 & 3.5 & 17.0 & 1.0 & 2.0 & 12.0 & 13.0 & 4.0 & 4.0 & 13.5 & 10.0 & 5.0 & 6.5 & 12.0 & 3.0 \\
 & \multicolumn{2}{l}{Avg. OOD Rank $\downarrow$} & 2.0 & 1.0 & 16.0 & 11.5 & 12.5 & 9.5 & 12.5 & 4.5 & 7 & 6 & 10 & 5 & 14.5 & 11.5 & 6.5 \\
\midrule
\multirow{6}{*}{\rotatebox[origin=c]{90}{Dim=4}}
  & \multicolumn{2}{l}{Complexity $\downarrow$}                   & \bestone{9.70} & \besttwo{10.60} & 14.40 & 23.10 & 18.50 & 20.60 & 19.60 & 17.90 & 18.70 & 14.50 & 13.90 & 13.90 & 13.60 & 22.60 & 22.10 \\
  & \multicolumn{2}{l}{Symbolic Prox. $\downarrow$}               & 32.70 & \bestone{30.30} & 36.40 & 34.10 & 32.90 & 34.70 & 32.70 & 32.50 & 32.40 & 34.80 & 32.90 & 32.20 & \besttwo{31.90} & 38.00 & 36.30 \\
  & \multirow{2}{*}{ID}  & $SR^2$                               & 11.50 & 74.60 & 34.80 & \besttwo{\underline{93.30}} &  \underline{91.00} & 49.40 & 61.10 & 78.00 & 78.70 & 88.60 & 86.90 & 86.10 & 89.10 & 78.70 & 80.00 \\
  &                       & $ACC_{0.9}$                           & 8.70 & 65.20 & 29.60 & \besttwo{\underline{93.90}} & \underline{89.30} & 40.70 & 52.00 & 74.10 & 75.00 & 78.80 & 70.00 & 71.40 & 85.70 & 74.10 & 80.80 \\
  & \multirow{2}{*}{OOD} & $SR^2$                               & 7.00 & 23.80 & 11.40 & 19.40 & {\besttwo{\underline{34.60}}} & 28.30 & 21.30 & \underline{30.70} & \bestone{\underline{36.90}} &  \underline{26.90} &  \underline{27.60} & 30.40 & 30.50 & 25.00 & 24.50 \\
  &                       & $ACC_{0.9}$                           & 0.00 & 21.70 & 7.10 & 12.50 & {\bestone{\underline{33.30}}} & 12.00 & 13.60 & 24.00 & \besttwo{30.80} & 19.40 & 20.70 & 15.40 & 21.40 & 21.70 & 17.40 \\
\cmidrule(lr){2-18}
 & \multicolumn{2}{l}{Avg. ID Rank $\downarrow$}  & 17.0 & 13.0 & 16.0 & 2.0 & 3.5 & 15.0 & 14.0 & 10.5 & 9.0 & 6.5 & 9.5 & 9.5 & 5.0 & 9.5 & 7.5 \\
 & \multicolumn{2}{l}{Avg. OOD Rank $\downarrow$} & 17.0 & 8.0 & 16.0 & 14.0 & 1.5 & 11.0 & 13.0 & 3.0 & 1.5 & 8.5 & 7.5 & 8.5 & 5.0 & 7.0 & 10.5 \\
\bottomrule
\end{tabular}
\end{threeparttable}
\end{table*}

\begin{figure}[t]
    \centering
    \includegraphics[width=0.8\linewidth]{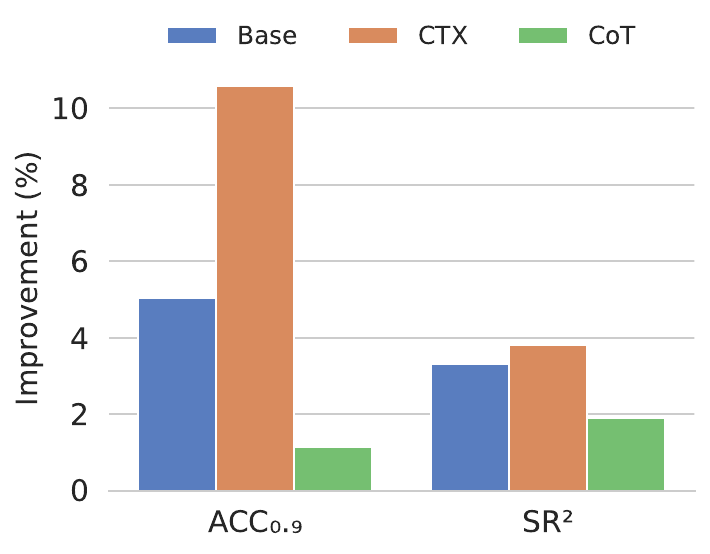}
    \vskip -1em
    \caption{Performance improvement of varying prompting strategies (Base, CTX, CoT) over the Na\"ive baseline. The context-aware strategy (CTX) consistently achieves the largest gains across metrics.}
    \label{fig: cde_improve}
    \vskip -1em
\end{figure}

\subsection{Experimental Setup}

\noindent\textbf{Baseline Models.}  
For CDEs, we include the following baselines: PySR~\cite{cranmer2024pysr}, a GP-based method; ProGED~\cite{omejc2024probabilistic}, a probabilistic grammar-based approach; and ODEformer~\cite{d2023odeformer}, a pretrained transformer-based model. For BNs, we evaluate LogicGep~\cite{zhang2024logicgep}, a GP-based method tailored for Boolean Network inference. For SCMs, we adopt several time-series causal discovery methods, including PCMCI~\cite{runge2019detecting}, LPCMCI~\cite{gerhardus2020high}, and j-PCMCI+~\cite{gunther2023causal}.


\noindent\textbf{Evaluation Metrics.}  
SymbolBench evaluates models on both in-distribution (ID) and out-of-distribution (OOD) data by applying new initial conditions. For \textbf{CDEs}, we report symbolic regression score: $SR^2 = \frac{1}{N} \sum_{i=1}^{N} R_i^2 \cdot \mathbb{I}(R_i^2 > 0)$, and accuracy for \( R^2 > 0.9 \): $ACC_{0.9}$. For \textbf{BNs}, we assess numerical similarity via precision, recall, F1, and bookmaker informedness: $recall + specificity - 1$. For \textbf{SCMs}, structural accuracy is measured using classification metrics and Structural Hamming Distance (SHD). Symbolic similarity (for CDEs and BNs) is evaluated via expression tree edit distance, and expression complexity via sympy’s \texttt{count\_ops}~\cite{10.7717/peerj-cs.103}. For BNs and SCMs, we also report accuracy for samples with F1 above a threshold: $ACC_{thesh}$.

\subsection{Experimental Results}
Across all three tasks, we select LLMs with various sizes and architectures, including Qwen2.5-14B~\cite{qwen2.5}, Llama-3.2-3B~\cite{dubey2024llama}, Mathstral-7B~\cite{jiang2023mistral7b}, GPT-4o-mini~\cite{achiam2023gpt}, and ChatTS-14B~\cite{xie2024chatts}. For GPT-4o-mini and ChatTS-14B, we use inputs with visual and temporal modalities. We present the results in Table~\ref{tab: CDEs}, ~\ref{tab: full_BN}, and ~\ref{tab: full_SCM}, where ablation studies on the four prompts with Qwen2.5-14B are integrated into the main tables rather than reported in a dedicated section.

\noindent\underline{\textbf{Obs. 1:}} \textbf{LLMs demonstrate superior capability compared to baselines on CDEs and SCMs datasets, while failing to compete against baselines in Boolean network inference.}
For Boolean network inference, while LLMs consistently achieve positive bookmaker informedness scores, indicating performance better than random guessing, the genetic programming-based model, LogicGep, significantly outperforms all evaluated LLMs across nearly all evaluation metrics in both ID and OOD scenarios.
This could be explained by both the symbolic and numerical fitting process. 
(a) \textit{Symbolic fitting} Unlike CDEs, BN inference does not involve coefficient optimization. Compared with CDEs, even if the symbolic structure is not precisely correct, adjusting coefficients can still yield good numerical accuracy. Compared with SCMs, SCM inference focuses only on discovering causal relationships, which is a less stringent goal than recovering the full dynamics as in BN inference. (b) \textit{Numerical fitting}: CDEs and SCMs are inferred from continuous time series, which form a consistent \textit{chains of state transitions}, whereas BN inference uses state transitions with various initial conditions, forming \textit{graphs of state transitions}, as shown in Appendix~\ref{appendix: dataset_examples} and ~\ref {appendix: pred_examples}, which can be hard to identify trends and summarize patterns. Further error analysis is provided in Appendix~\ref{appendix: BN_error}.



\noindent\underline{\textbf{Obs. 2:}} 
\textbf{LLMs' performance degrades with problem difficulty.} Across all models, performance consistently declines as the dimensionality of the system increases. While conventional methods such as PySR remain competitive in low-dimensional systems (e.g., dim = 1) in Table~\ref{tab: CDEs}, baseline models tend to exhibit a steeper performance drop compared to LLMs on higher dimensional systems (full results in Appendix~\ref{appendix: full_bn_scm}). Similarly, there is a notable drop in accuracy when models are evaluated on OOD data, underscoring the increased complexity and generalization challenges posed by these scenarios.

\noindent\underline{\textbf{Obs. 3:}} \textbf{Chain-of-thought prompting does not consistently improve performance.} Except ChatTS-14B, further introducing CoT prompting does not lead to consistent gains, especially on the CDEs dataset, as shown in Figure~\ref{fig: cde_improve}. While increasing test-time compute has been shown to improve outcomes in various tasks~\cite{snell2024scaling}, to which CoT can contribute, similar limitations are observed in other benchmarks, e.g., SciBench~\cite{wang2023scibench}.

\noindent\underline{\textbf{Obs. 4:}} \textbf{Providing problem contexts improves LLM performance.} Across all three tasks, compared to the Naive prompt, LLMs demonstrate the ability to leverage the provided context, resulting in higher numerical performance. In this setting, context functions as a form of conditioning, helping to constrain the solution space and guide the model toward more accurate and relevant inferences with a faster convergence rate, as shown in Appendix~\ref{appendix: converge_rate}.

\begin{table*}[t]
\centering
\caption{Comparison of Boolean network inference across ID and OOD settings. Full results in Appendix~\ref{appendix: full_bn_scm}. \textcolor{orange}{Orange} and \textcolor{yellow}{Yellow} mark the first and second place, respectively.}
\vskip -1em
\label{tab: full_BN}
\setlength{\tabcolsep}{2pt}
\renewcommand{\arraystretch}{1.1}
\resizebox{\textwidth}{!}{%
\begin{tabular}{c c ccccccc ccccccc c c}
\toprule
\multirow{2}{*}{\textbf{Model}} & \multirow{2}{*}{\textbf{Setting}}
& \multicolumn{7}{c}{\textbf{ID}}
& \multicolumn{7}{c}{\textbf{OOD}}
& \multirow{2}{*}{\textbf{Symb. Prox.} $\downarrow$}
& \multirow{2}{*}{\textbf{Comp.} $\downarrow$} \\
\cmidrule(lr){3-9}\cmidrule(lr){10-16}
& & \textbf{Prec.} & \textbf{Rec.} & \textbf{Acc.} & \textbf{B.I.} & $\mathbf{ACC_{0.5}}$ & $\mathbf{ACC_{0.7}}$ & $\mathbf{ACC_{0.8}}$
  & \textbf{Prec.} & \textbf{Rec.} & \textbf{Acc.} & \textbf{B.I.} & $\mathbf{ACC_{0.5}}$ & $\mathbf{ACC_{0.7}}$ & $\mathbf{ACC_{0.8}}$
  & & \\
\midrule
\multirow{1}{*}{LogicGep}
& $\sim$         & \bestone{93.6} & \bestone{92.7} & \bestone{95.2} & \bestone{88.7} & \bestone{98.5} & \bestone{98.5} & \bestone{96.9}
                 & \bestone{84.7} & \bestone{86.5} & \bestone{89.5} & \bestone{76.5} & \bestone{98.5} & \bestone{86.2} & \bestone{76.9}
                 & 12.39 & \bestone{12.39} \\
\midrule
\multirow{4}{*}{Qwen2.5-14B}
& Na\"ive        & \besttwo{58.7} & 71.7 & 66.5 & 29.5 & 86.2 & 33.8 & 6.2
                 & \besttwo{56.4} & 71.3 & 64.5 & 26.6 & 80.0 & 24.6 & 7.7
                 & 12.87 & 14.84 \\
& Base           & 54.7 & 73.3 & 63.5 & 27.3 & 80.0 & 20.0 & 3.1
                 & 53.8 & 72.8 & 62.5 & 25.9 & 83.1 & 16.9 & 3.1
                 & \besttwo{12.33} & 16.73 \\
& Context        & 57.8 & \besttwo{77.2} & \besttwo{67.1} & 29.1 & 87.7 & \besttwo{38.5} & \besttwo{9.2}
                 & 56.1 & \besttwo{77.2} & \besttwo{65.5} & 27.0 & \besttwo{87.7} & \besttwo{30.8} & \besttwo{10.8}
                 & 12.39 & 16.86 \\
& CoT            & 58.3 & 73.9 & 65.4 & \besttwo{30.3} & \besttwo{92.3} & 27.7 & 4.6
                 & 56.2 & 73.9 & 63.2 & \besttwo{27.9} & 84.6 & 24.6 & 3.1
                 & \bestone{11.96} & \besttwo{14.79} \\
\bottomrule
\end{tabular}}
\end{table*}


\begin{table}[t]
\centering
\caption{Comparison of causal discovery methods and LLM-based approaches. Full results in Appendix ~\ref{appendix: full_bn_scm}.}
\vskip -0.8em
\label{tab: full_SCM}
\setlength{\tabcolsep}{1.2pt}
\renewcommand{\arraystretch}{0.8}
\resizebox{0.45\textwidth}{!}{%
\begin{tabular}{c c c c c c c c}
\toprule
\textbf{Model} & \textbf{Setting} & \textbf{F1} & \textbf{FDR$\downarrow$} & $\mathbf{ACC_{0.7}}$ & $\mathbf{ACC_{0.8}}$ & \textbf{SHD$\downarrow$} \\
\midrule
PCMCI          & $\sim$           & 52.7 & 46.1 & \besttwo{23.7} & 7.9  & 95.28 \\
LPCMCI         & $\sim$           & 52.0 & \besttwo{29.1} & 18.4 & 5.3  & \bestone{25.35} \\
j-PCMCI+       & $\sim$           & 46.2 & 39.1 & 13.2 & 7.4  & 49.36 \\
\midrule
\multirow{4}{*}{Qwen2.5-14B}
               & Na\"ive          & 49.7 & 40.6 & 17.9 & 8.4  & 35.54 \\
               & Base             & 50.5 & 38.6 & 18.4 & 8.9  & 34.58 \\
               & Context          & 53.4 & 37.0 & 19.5 & 7.9  & 31.69 \\
               & CoT              & 51.3 & 38.1 & 20.0 & 8.4  & 39.36 \\
\midrule
\multirow{2}{*}{ChatTS-14B}
               & Context          & \besttwo{54.1} & \bestone{27.7} & 22.6 & \bestone{11.6} & 32.48 \\
               & CoT              & \bestone{54.4} & \bestone{27.7} & \bestone{25.3} & \besttwo{10.5} & \besttwo{31.10} \\
\bottomrule
\end{tabular}}
\vskip -0.8em
\end{table}

\begin{table}[h]
\centering
\caption{Correlations between complexity and OOD $ACC_{0.9}$ across four dimensions with and without context in symbolic regression task.}
\label{tab: CDE_corr}
\vskip -0.8em
\resizebox{0.4\textwidth}{!}{%
\begin{tabular}{l
                S[table-format=+1.3]
                S[table-format=+1.3]
                S[table-format=+1.3]
                S[table-format=+1.3]}
\toprule
\textbf{Condition} & \textbf{Dim 1} & \textbf{Dim 2} & \textbf{Dim 3} & \textbf{Dim 4} \\
\midrule
w/o context & -0.672 & -0.384 & -0.425 &  0.584 \\
w/  context &  0.097 &  0.728 &  0.165 & -0.174 \\
\bottomrule
\end{tabular}}
\vskip -1em
\end{table}

\subsection{Further Analysis}
In this section, we explore the adaptability of our framework with genetic programming methods and the power of test-time compute. We formulate several key questions and takeaways as follows: 

\noindent\textbf{Q1: Do LLMs Employ Correct Symbolic Reasoning Paths?}
While previous research~\cite{hao2024llm} has demonstrated that yielding correct answers does not necessarily yield correct reasoning paths, we find that general LLMs like Qwen2.5-14B with pure textual input are able to perform a certain level of reasoning (through CoT prompting) over the time series and the given context. As shown in Table~\ref{appendix: CoT_example}, the LLM is not only able to consider the meaning of each variable, but also the historical candidates.

\noindent\textbf{Q2: How should test-time compute be structured for consistent improvement?}
While scaling test-time compute can boost LLM performance~\cite{snell2024scaling}, our results show that the structure of scaling is crucial for consistent gains. Simply applying naive CoT reasoning offers limited benefit (\textit{Obs. 3}) due to shallow reasoning depth and the absence of mechanisms like reflection and verification. In contrast, advanced \textbf{RLM}~\cite{besta2025reasoning} with \textbf{Long CoT}~\cite{chen2025towards} incorporates these cognitive processes, leading to more robust improvements. Illustrative examples are provided in Appendix~\ref{appendix: RLM_example}.

\begin{figure}
    \centering
    \includegraphics[width=\linewidth]{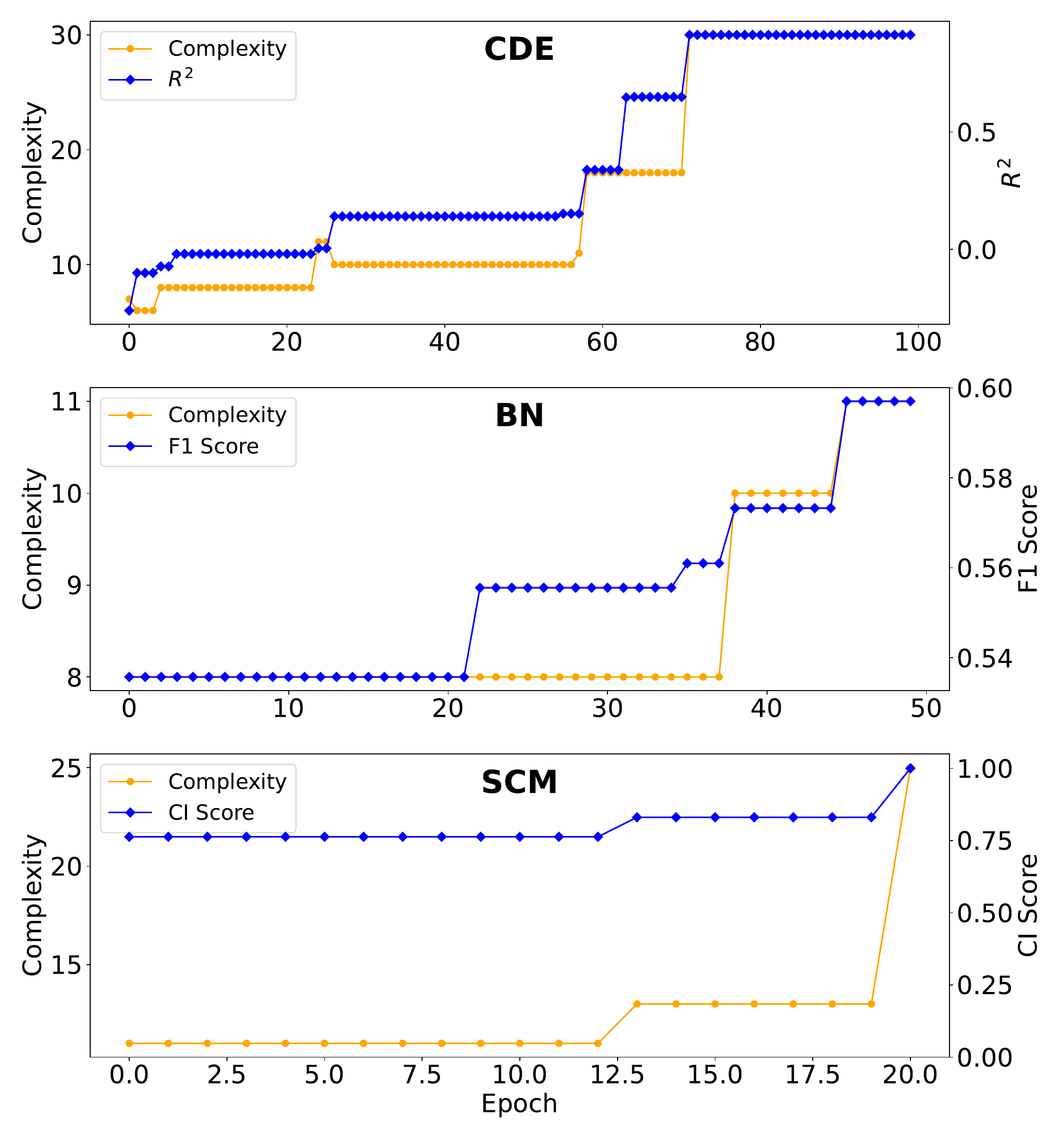}
    \vskip -1em
    \caption{Evaluation scores improve with more iterations and test-time compute, as discussed in \textbf{Q2}.}
    \label{fig: DE_epoch_scores}
\end{figure}

\begin{figure}[t]
    \centering
    \begin{subfigure}[b]{0.45\textwidth}
        \centering
        \includegraphics[width=\linewidth]{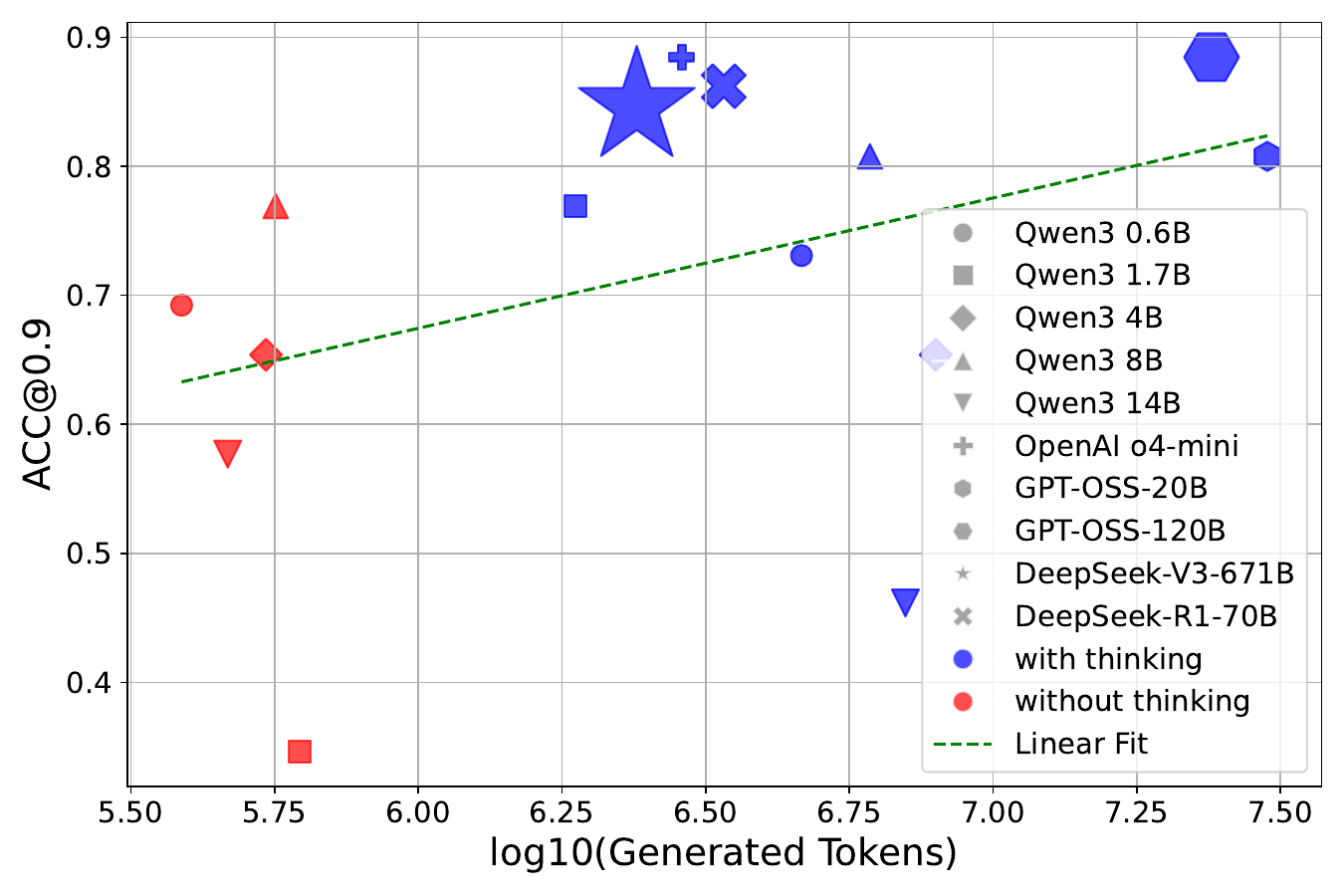}
        \vskip -0.5em
        \caption{Comparison of increasing test-time compute on CDEs.}
        \label{fig: DE_compute}
        \vskip -0.5em
    \end{subfigure}
    \hfill
    \begin{subfigure}[b]{0.48\textwidth}
        \centering
        \includegraphics[width=\linewidth]{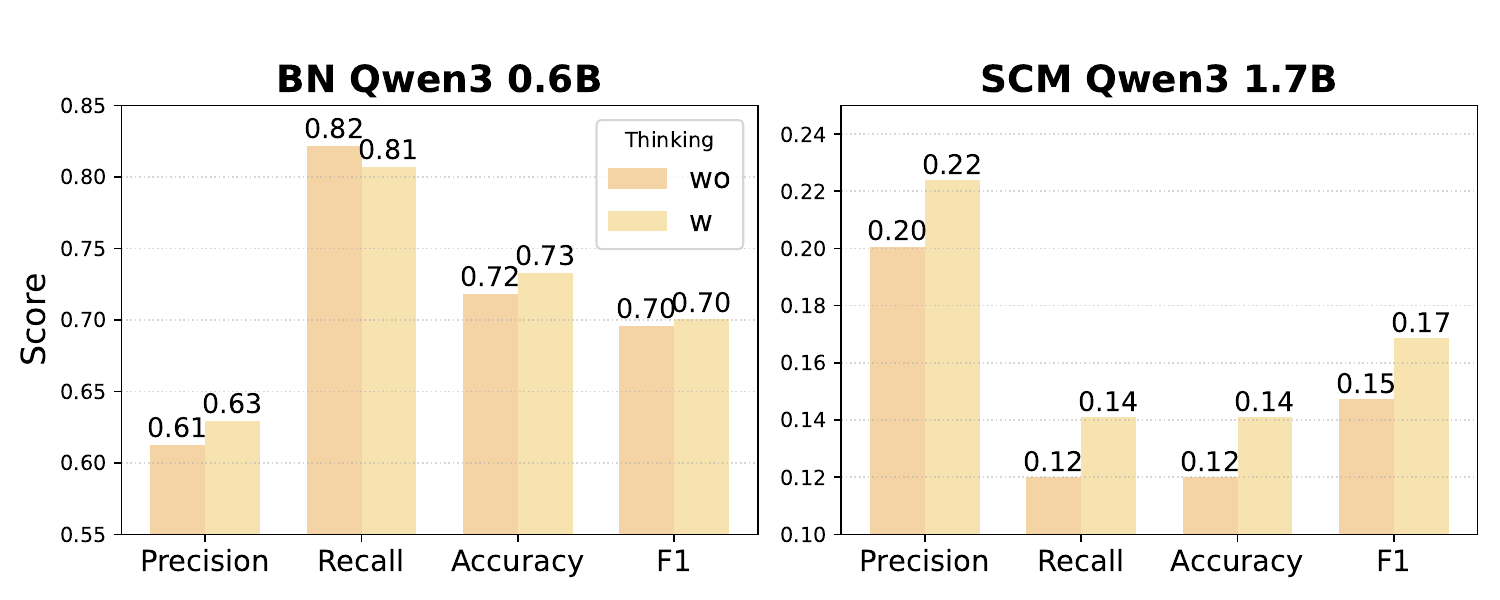}
        \vskip -0.5em
        \caption{Comparison of w and w/o thinking on BNs and SCMs.}
        \label{fig: BN_SCM_compute}
        
    \end{subfigure}
    \vskip -0.5em
    \caption{Comparison of performance on thinking.}
    \label{fig: compute_comparison}
\end{figure}

Thus, to see consistent gains, test-time compute should be structured to facilitate a Long CoT. We explore two approaches: \textbf{First}, we analyze the effect of a Long CoT, generated by RLMs, within each reasoning epoch. This reveals that dedicating more compute to a longer, more detailed chain-of-thought leads to moderate but consistent improvements in the final prediction scores, as shown in Figure~\ref{fig: compute_comparison}. \textbf{Second}, we view the entire iterative refinement process as a form of Long CoT. In this approach, the model verifies previous outputs and generates improved answers based on earlier, potentially flawed, solutions. As shown in Figure~\ref{fig: DE_epoch_scores}, the verification scores consistently improve as more refinement steps are added. From both approaches, we observe that the complexity of the best-fitting solutions increases with the amount of computation, echoing the human-like process of progressively constructing more sophisticated answers.

\begin{table}[t]
\centering
\caption{Hybrid method on CDEs. We use the same GPT-4o-mini as judge and predictor.}
\label{tab: hybrid_CDE}
\resizebox{0.46\textwidth}{!}{%
\begin{tabular}{lcccc}
\toprule
\textbf{Method} &
\textbf{$SR_{ID}^2$} &
\textbf{$SR_{OOD}^2$} &
\textbf{Sym.\ Prox.\,$\downarrow$} &
\textbf{Comp.\,$\downarrow$} \\
\midrule
GPLearn                                                          & 27.6 & 14.9 & 6.083 & 1.333 \\
GPT-4o-mini                                                      & 39.3 & 24.1 & 7.438 & 7.406 \\ \midrule
\shortstack[l]{GPLearn +\\ LLM-as-Judge}                         & 31.7 & 16.3 & 6.550 & \textbf{1.200} \\
\shortstack[l]{GPLearn + \\ LLM-as-Predictor}                     & \textbf{89.5} & \textbf{69.3} & \textbf{5.045} & 2.682 \\
\bottomrule
\end{tabular}}
\vskip -1em
\end{table}

\begin{table}[h]
\centering
\caption{Hybrid method on BNs. We use the same GPT-4o-mini as judge and predictor.}
\label{tab: hybrid_BN}
\resizebox{0.47\textwidth}{!}{%
\begin{tabular}{llccccc c}
\toprule
\multirow{2}{*}{\textbf{Method}} & \multirow{2}{*}{\textbf{Setting}} &
\multicolumn{5}{c}{\textbf{Metrics}} & \textbf{Comp.\,$\downarrow$} \\
\cmidrule(lr){3-7}
 & & \textbf{Prec.} & \textbf{Recall} & \textbf{F1} & \textbf{Acc} & \textbf{BM} &  \\
\midrule
\multirow{2}{*}{LogicGep} & ID  & 88.0 & 88.6 & 88.0 & 91.7 & 80.9 & \multirow{2}{*}{\textbf{0.773}} \\
 & OOD & \textbf{78.9} & 83.0 & 80.0 & 85.6 & 69.2 &  \\
\multirow{2}{*}{GPT-4o-mini} & ID & 52.8 & 65.6 & 57.8 & 61.2 & 23.6 & \multirow{2}{*}{2.422} \\
 & OOD & 49.9 & 62.3 & 54.5 & 58.3 & 17.3 &  \\
 \midrule
\multirow{2}{*}{\shortstack[l]{LogicGep +\\ Judge}} 
 & ID & \textbf{89.3} & \textbf{92.8} & \textbf{90.9} & \textbf{93.2} & \textbf{85.6} & \multirow{2}{*}{1.012} \\
 & OOD & 78.6 & \textbf{85.2} & \textbf{80.4} & \textbf{85.8} & \textbf{71.0} &  \\
\multirow{2}{*}{\shortstack[l]{Predictor +\\ LogicGep}} 
 & ID & 72.2 & 84.7 & 77.5 & 80.6 & 61.0 & \multirow{2}{*}{0.993} \\
 & OOD & 63.1 & 77.2 & 67.9 & 73.2 & 46.9 &  \\
\bottomrule
\end{tabular}}
\vskip -0.8em
\end{table}

\begin{figure}[ht]
    \centering
    \includegraphics[width=\linewidth]{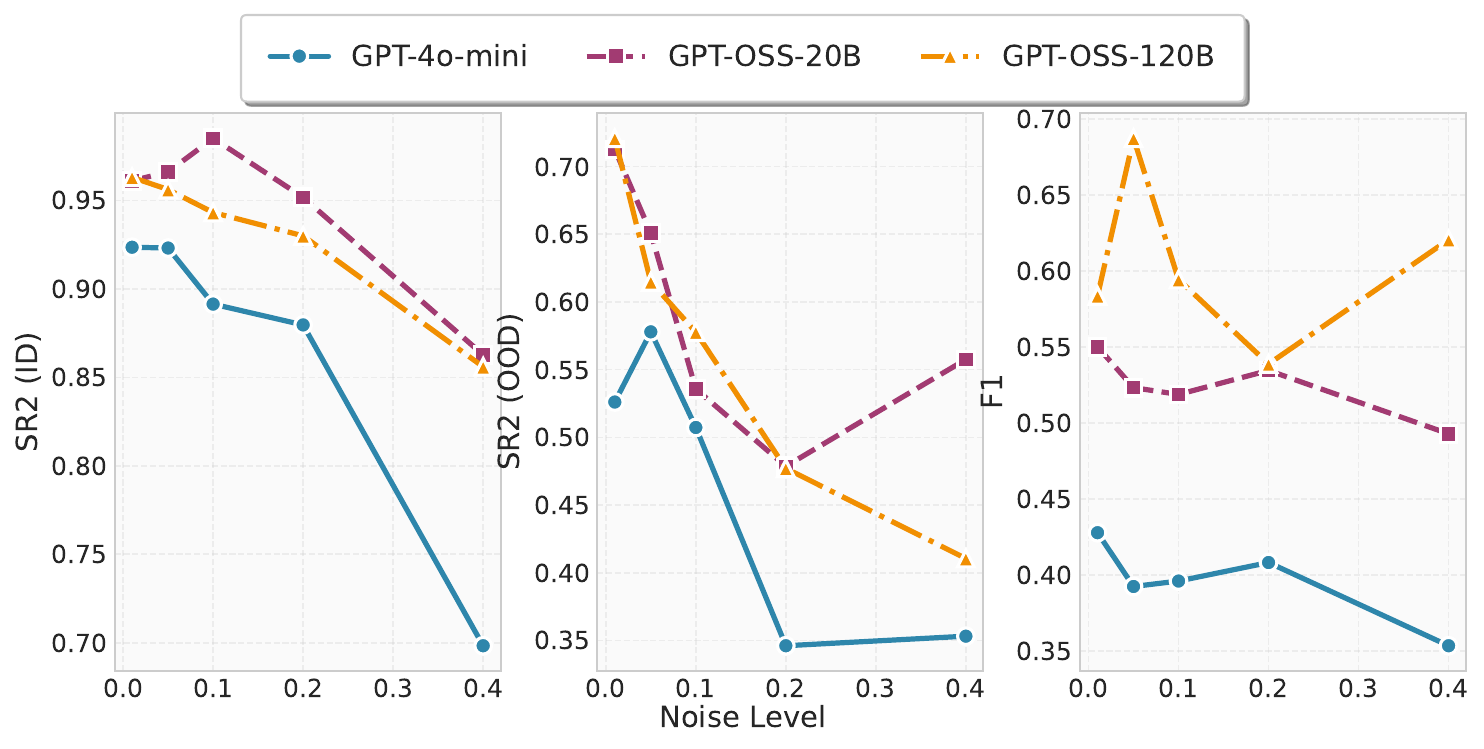}
    \caption{Performance degradation on the CDE (Left, Middle) and SCM (Right) datasets after introducing Gaussian noise to the time series.}
    \label{fig: noise_DE_SCM}
\end{figure}


\noindent\textbf{Q3: Does the generalizability of prediction correlate with structural complexity?}
When multiple symbolic structures fit a time series, conventional wisdom, rooted in Occam’s Razor, favors the simplest as most generalizable. Recent LLM-based symbolic regression methods~\cite{shojaee2024llm, li2024mllm, merler2024context, grayeli2024symbolic, wang2025drsr} follow this principle. However, our results challenge the notion that simpler is always better. As shown in Table~\ref{tab: full-symbolic}, more complex expressions can yield higher symbolic proximity and OOD performance. In domains like biology, physics, and healthcare~\cite{xie2025kerap}, true underlying laws may be structurally complex yet numerically simple in-distribution. Overemphasizing simplicity can therefore obscure more generalizable, \textit{context-aligned} candidates. As detailed in Appendix~\ref{appendix: corr_CDE} and Table~\ref{tab: CDE_corr}, our findings indicate that: \textit{\uline{Candidate ranking and selection should not solely rely on complexity and may benefit from considering contextual information}}.

\noindent\textbf{Q4: How do LLMs perform with noisy data?}
To evaluate robustness to real-world data imperfections, we introduce Gaussian noise at varying levels (0.01 to 0.4) to the time series data and assess performance degradation across different model scales. We sample 20 cases for each experiment. As shown in Figure~\ref{fig: noise_DE_SCM}, on the CDE dataset, most models remain vulnerable to the increasing noise, especially in the OOD setting. On the other hand, on the SCM dataset, larger models exhibit increasing robustness. Overall, all models experience performance degradation, and the impact varies across tasks.

\noindent\textbf{Q5: Can LLMs and genetic programming (GP) be combined for enhanced performance?} While LLMs show superior performance on CDE and SCM inference, GP holds value for its cost, control over complexity, and stronger performance in BN inference. To capitalize on the advantages of both, we explored two primary methods for creating a high-performance hybrid system, as detailed in Figure~\ref{fig: hybrid_method_} and Appendix~\ref{appendix: hybrid_method}. The key is to use the contextual understanding of LLMs to ``inject context and knowledge'' into the GP workflow, boosting its performance. This can be achieved in two main ways: (a) \textit{LLM-Guided Initialization}: Use an "LLM-as-Predictor" to generate a high-quality initial population of candidate expressions for the GP to evolve. This provides the GP with a context-aware and promising starting point. (b) \textit{LLM-Guided Evaluation}: Employ an ``LLM-as-Judge'' to provide context-enhanced evaluation scores during the GP's fitness assessment. This helps guide the evolutionary search toward solutions that are not only numerically accurate but also scientifically plausible. As shown in Tables~\ref{tab: hybrid_CDE} and \ref{tab: hybrid_BN}, combining LLMs with GP in these roles demonstrably improves performance.

\section{Conclusion and Outlook}
We introduce \textbf{SymbolBench}, a real-world benchmark for symbolic structure discovery, and a \textbf{Unified Symbolic Reasoning Framework} that enables LLMs (optionally with GP) to generate and judge hypotheses across tasks. Experiments show: (i) LLMs beat baselines on multivariate symbolic regression and causal discovery but lag on Boolean network inference; (ii) more test-time compute yields only modest gains; (iii) contextual grounding boosts accuracy and generalizability; and (iv) LLM–GP hybrids further improve performance.

\vskip 0.2em
\noindent\textbf{Future Opportunities.} 
{Scientific discovery from observations lies at the core of human science development. Focusing on one specific type of observation: temporal data, we highlight several opportunities to push the development of automation in scientific discovery: (i) task-specific scaling of test-time compute and reasoning depth; (ii) knowledge-heavy \textbf{context} and context-aware \textbf{verifications} beyond syntactic simplicity to guide hypotheses; (iii) adaptation to diverse multimodal observations, and (iv) integration with efficient Genetic Programming algorithms.}

\clearpage

\section{Limitations}
This work has several limitations stemming from computational constraints. First, we provide evaluations on large open-source models (e.g., DeepSeek-R1) only in Figure~\ref{fig: compute_comparison} instead of in the main tables. Second, we capped the budget at 100 generation epochs and a maximum of 20 retries per epoch for each run; samples without candidates that could reach the tolerance will be rerun once again to mitigate uncertainty. Finally, we examined only one LLM and one GP in the hybrid architecture, and did not analyze the effects of reasoning strategies or model size.

\section*{Acknowledgments}
Zewen Liu and Wei Jin are supported by the U.S. National Science Foundation under Award Number 2504088, and the National Institute of Allergy and Infectious Diseases of the NIH under Award Numbers R01AI189829 and R01AI197111. The content is the sole responsibility of the authors and does not necessarily represent the views of the NIH. This project is also partially supported by the NAIRR Pilot Program.

\bibliography{custom}

\begin{thebibliography}{49}
\providecommand{\natexlab}[1]{#1}

\bibitem[{Achiam et~al.(2023)Achiam, Adler, Agarwal, Ahmad, Akkaya, Aleman, Almeida, Altenschmidt, Altman, Anadkat et~al.}]{achiam2023gpt}
Josh Achiam, Steven Adler, Sandhini Agarwal, Lama Ahmad, Ilge Akkaya, Florencia~Leoni Aleman, Diogo Almeida, Janko Altenschmidt, Sam Altman, Shyamal Anadkat, and 1 others. 2023.
\newblock Gpt-4 technical report.
\newblock \emph{arXiv preprint arXiv:2303.08774}.

\bibitem[{Ahn et~al.(2024)Ahn, Verma, Lou, Liu, Zhang, and Yin}]{ahn2024large}
Janice Ahn, Rishu Verma, Renze Lou, Di~Liu, Rui Zhang, and Wenpeng Yin. 2024.
\newblock \href {https://aclanthology.org/2024.eacl-srw.17} {Large language models for mathematical reasoning: Progresses and challenges}.
\newblock In \emph{Proceedings of the 18th Conference of the European Chapter of the Association for Computational Linguistics, {EACL} 2024: Student Research Workshop, St. Julian's, Malta, March 21-22, 2024}, pages 225--237. Association for Computational Linguistics.

\bibitem[{Assaad et~al.(2022)Assaad, Devijver, and Gaussier}]{assaad2022survey}
Charles~K Assaad, Emilie Devijver, and Eric Gaussier. 2022.
\newblock Survey and evaluation of causal discovery methods for time series.
\newblock \emph{Journal of Artificial Intelligence Research}, 73:767--819.

\bibitem[{Besta et~al.(2025)Besta, Barth, Schreiber, Kubicek, Catarino, Gerstenberger, Nyczyk, Iff, Li, Houliston et~al.}]{besta2025reasoning}
Maciej Besta, Julia Barth, Eric Schreiber, Ales Kubicek, Afonso Catarino, Robert Gerstenberger, Piotr Nyczyk, Patrick Iff, Yueling Li, Sam Houliston, and 1 others. 2025.
\newblock Reasoning language models: A blueprint.
\newblock \emph{arXiv preprint arXiv:2501.11223}.

\bibitem[{Brunton et~al.(2016)Brunton, Proctor, and Kutz}]{brunton2016discovering}
Steven~L Brunton, Joshua~L Proctor, and J~Nathan Kutz. 2016.
\newblock Discovering governing equations from data by sparse identification of nonlinear dynamical systems.
\newblock \emph{Proceedings of the national academy of sciences}, 113(15):3932--3937.

\bibitem[{Carson et~al.(2024)Carson, Chen, and Kang}]{carson2024llm}
Erin Carson, Xinye Chen, and Cheng Kang. 2024.
\newblock Llm-abba: Understanding time series via symbolic approximation.
\newblock \emph{arXiv preprint arXiv:2411.18506}.

\bibitem[{Chen et~al.(2025)Chen, Qin, Liu, Peng, Guan, Wang, Hu, Zhou, Gao, and Che}]{chen2025towards}
Qiguang Chen, Libo Qin, Jinhao Liu, Dengyun Peng, Jiannan Guan, Peng Wang, Mengkang Hu, Yuhang Zhou, Te~Gao, and Wanxiang Che. 2025.
\newblock Towards reasoning era: A survey of long chain-of-thought for reasoning large language models.
\newblock \emph{arXiv preprint arXiv:2503.09567}.

\bibitem[{Cranmer(2024)}]{cranmer2024pysr}
Miles Cranmer. 2024.
\newblock Pysr: High-performance symbolic regression in python and julia.
\newblock \emph{Astrophysics Source Code Library}, pages ascl--2409.

\bibitem[{d'Ascoli et~al.(2023)d'Ascoli, Becker, Mathis, Schwaller, and Kilbertus}]{d2023odeformer}
St{\'e}phane d'Ascoli, S{\"o}ren Becker, Alexander Mathis, Philippe Schwaller, and Niki Kilbertus. 2023.
\newblock Odeformer: Symbolic regression of dynamical systems with transformers.
\newblock \emph{arXiv preprint arXiv:2310.05573}.

\bibitem[{Dubey et~al.(2024)Dubey, Jauhri, Pandey, Kadian, Al-Dahle, Letman, Mathur, Schelten, Yang, Fan et~al.}]{dubey2024llama}
Abhimanyu Dubey, Abhinav Jauhri, Abhinav Pandey, Abhishek Kadian, Ahmad Al-Dahle, Aiesha Letman, Akhil Mathur, Alan Schelten, Amy Yang, Angela Fan, and 1 others. 2024.
\newblock The llama 3 herd of models.
\newblock \emph{arXiv e-prints}, pages arXiv--2407.

\bibitem[{Fang et~al.(2024)Fang, Deng, Zhang, Shi, Chen, Pechenizkiy, and Wang}]{fang2024large}
Meng Fang, Shilong Deng, Yudi Zhang, Zijing Shi, Ling Chen, Mykola Pechenizkiy, and Jun Wang. 2024.
\newblock Large language models are neurosymbolic reasoners.
\newblock In \emph{Proceedings of the AAAI conference on artificial intelligence}, volume~38, pages 17985--17993.

\bibitem[{Gao et~al.(2022)Gao, Sun, Xiang, Qin, and Lee}]{gao2022LearningAsynchronousBoolean}
Shuhua Gao, Changkai Sun, Cheng Xiang, Kairong Qin, and Tong~Heng Lee. 2022.
\newblock \href {https://doi.org/10.1109/TCYB.2020.3022430} {Learning {{Asynchronous Boolean Networks From Single-Cell Data Using Multiobjective Cooperative Genetic Programming}}}.
\newblock \emph{IEEE Transactions on Cybernetics}, 52(5):2916--2930.

\bibitem[{Gentner(2002)}]{gentner2002analogy}
Dedre Gentner. 2002.
\newblock Analogy in scientific discovery: The case of johannes kepler.
\newblock In \emph{Model-based reasoning: Science, technology, values}, pages 21--39. Springer.

\bibitem[{Gerhardus and Runge(2020)}]{gerhardus2020high}
Andreas Gerhardus and Jakob Runge. 2020.
\newblock High-recall causal discovery for autocorrelated time series with latent confounders.
\newblock \emph{Advances in neural information processing systems}, 33:12615--12625.

\bibitem[{Glenn et~al.(2024)Glenn, Dakle, Wang, and Raghavan}]{glenn2024blendsql}
Parker Glenn, Parag Dakle, Liang Wang, and Preethi Raghavan. 2024.
\newblock Blendsql: A scalable dialect for unifying hybrid question answering in relational algebra.
\newblock In \emph{Findings of the Association for Computational Linguistics: ACL 2024}, pages 453--466.

\bibitem[{Grayeli et~al.(2024)Grayeli, Sehgal, Costilla~Reyes, Cranmer, and Chaudhuri}]{grayeli2024symbolic}
Arya Grayeli, Atharva Sehgal, Omar Costilla~Reyes, Miles Cranmer, and Swarat Chaudhuri. 2024.
\newblock Symbolic regression with a learned concept library.
\newblock \emph{Advances in Neural Information Processing Systems}, 37:44678--44709.

\bibitem[{G{\"u}nther et~al.(2023)G{\"u}nther, Ninad, and Runge}]{gunther2023causal}
Wiebke G{\"u}nther, Urmi Ninad, and Jakob Runge. 2023.
\newblock Causal discovery for time series from multiple datasets with latent contexts.
\newblock In \emph{Uncertainty in Artificial Intelligence}, pages 766--776. PMLR.

\bibitem[{Hao et~al.(2024)Hao, Gu, Luo, Liu, Shao, Wang, Xie, Ma, Samavedhi, Gao et~al.}]{hao2024llm}
Shibo Hao, Yi~Gu, Haotian Luo, Tianyang Liu, Xiyan Shao, Xinyuan Wang, Shuhua Xie, Haodi Ma, Adithya Samavedhi, Qiyue Gao, and 1 others. 2024.
\newblock Llm reasoners: New evaluation, library, and analysis of step-by-step reasoning with large language models.
\newblock \emph{arXiv preprint arXiv:2404.05221}.

\bibitem[{Hasan et~al.(2023)Hasan, Hossain, and Gani}]{hasan2023survey}
Uzma Hasan, Emam Hossain, and Md~Osman Gani. 2023.
\newblock A survey on causal discovery methods for iid and time series data.
\newblock \emph{arXiv preprint arXiv:2303.15027}.

\bibitem[{Jiang et~al.(2023)Jiang, Sablayrolles, Mensch, Bamford, Chaplot, de~las Casas, Bressand, Lengyel, Lample, Saulnier, Lavaud, Lachaux, Stock, Scao, Lavril, Wang, Lacroix, and Sayed}]{jiang2023mistral7b}
Albert~Q. Jiang, Alexandre Sablayrolles, Arthur Mensch, Chris Bamford, Devendra~Singh Chaplot, Diego de~las Casas, Florian Bressand, Gianna Lengyel, Guillaume Lample, Lucile Saulnier, Lélio~Renard Lavaud, Marie-Anne Lachaux, Pierre Stock, Teven~Le Scao, Thibaut Lavril, Thomas Wang, Timothée Lacroix, and William~El Sayed. 2023.
\newblock \href {https://arxiv.org/abs/2310.06825} {Mistral 7b}.
\newblock \emph{Preprint}, arXiv:2310.06825.

\bibitem[{Lei et~al.(2025)Lei, Han, Rossi, Dernoncourt, Lipka, Halappanavar, Tang, and Wang}]{lei2025mixture}
Yongjia Lei, Haoyu Han, Ryan~A Rossi, Franck Dernoncourt, Nedim Lipka, Mahantesh~M Halappanavar, Jiliang Tang, and Yu~Wang. 2025.
\newblock Mixture of structural-and-textual retrieval over text-rich graph knowledge bases.
\newblock In \emph{Findings of the Association for Computational Linguistics: ACL 2025}, pages 18306--18321.

\bibitem[{Li et~al.(2024)Li, Li, Yu, Wu, Liu, Li, Wei, and Deng}]{li2024mllm}
Yanjie Li, Weijun Li, Lina Yu, Min Wu, Jingyi Liu, Wenqiang Li, Shu Wei, and Yusong Deng. 2024.
\newblock Mllm-sr: Conversational symbolic regression base multi-modal large language models.
\newblock \emph{arXiv preprint arXiv:2406.05410}.

\bibitem[{Li et~al.(2023)Li, Li, and Yan}]{li2023time}
Zekun Li, Shiyang Li, and Xifeng Yan. 2023.
\newblock Time series as images: Vision transformer for irregularly sampled time series.
\newblock \emph{Advances in Neural Information Processing Systems}, 36:49187--49204.

\bibitem[{Liu et~al.(2025)Liu, Chung, Chen, Yeung, and Imani}]{liu2025hypervectors}
Yezi Liu, William~Youngwoo Chung, Hanning Chen, Calvin Yeung, and Mohsen Imani. 2025.
\newblock Are hypervectors enough? single-call llm reasoning over knowledge graphs.
\newblock \emph{arXiv preprint arXiv:2512.09369}.

\bibitem[{Makke and Chawla(2024)}]{makke2024interpretable}
Nour Makke and Sanjay Chawla. 2024.
\newblock Interpretable scientific discovery with symbolic regression: a review.
\newblock \emph{Artificial Intelligence Review}, 57(1):2.

\bibitem[{Merler et~al.(2024)Merler, Haitsiukevich, Dainese, and Marttinen}]{merler2024context}
Matteo Merler, Katsiaryna Haitsiukevich, Nicola Dainese, and Pekka Marttinen. 2024.
\newblock In-context symbolic regression: Leveraging large language models for function discovery.
\newblock \emph{arXiv preprint arXiv:2404.19094}.

\bibitem[{Meurer et~al.(2017)Meurer, Smith, Paprocki, \v{C}ert\'{i}k, Kirpichev, Rocklin, Kumar, Ivanov, Moore, Singh, Rathnayake, Vig, Granger, Muller, Bonazzi, Gupta, Vats, Johansson, Pedregosa, Curry, Terrel, Rou\v{c}ka, Saboo, Fernando, Kulal, Cimrman, and Scopatz}]{10.7717/peerj-cs.103}
Aaron Meurer, Christopher~P. Smith, Mateusz Paprocki, Ond\v{r}ej \v{C}ert\'{i}k, Sergey~B. Kirpichev, Matthew Rocklin, AMiT Kumar, Sergiu Ivanov, Jason~K. Moore, Sartaj Singh, Thilina Rathnayake, Sean Vig, Brian~E. Granger, Richard~P. Muller, Francesco Bonazzi, Harsh Gupta, Shivam Vats, Fredrik Johansson, Fabian Pedregosa, and 8 others. 2017.
\newblock \href {https://doi.org/10.7717/peerj-cs.103} {Sympy: symbolic computing in python}.
\newblock \emph{PeerJ Computer Science}, 3:e103.

\bibitem[{Mundhenk et~al.(2021)Mundhenk, Landajuela, Glatt, Santiago, Faissol, and Petersen}]{mundhenk2021symbolic}
T~Nathan Mundhenk, Mikel Landajuela, Ruben Glatt, Claudio~P Santiago, Daniel~M Faissol, and Brenden~K Petersen. 2021.
\newblock Symbolic regression via neural-guided genetic programming population seeding.
\newblock \emph{arXiv preprint arXiv:2111.00053}.

\bibitem[{Omejc et~al.(2024)Omejc, Gec, Brence, Todorovski, and D{\v{z}}eroski}]{omejc2024probabilistic}
Nina Omejc, Bo{\v{s}}tjan Gec, Jure Brence, Ljup{\v{c}}o Todorovski, and Sa{\v{s}}o D{\v{z}}eroski. 2024.
\newblock Probabilistic grammars for modeling dynamical systems from coarse, noisy, and partial data.
\newblock \emph{Machine learning}, 113(10):7689--7721.

\bibitem[{Pastva et~al.(2023)Pastva, {\v{S}}afr{\'a}nek, Bene{\v{s}}, Brim, and Henzinger}]{pastva2023repository}
Samuel Pastva, David {\v{S}}afr{\'a}nek, Nikola Bene{\v{s}}, Lubo{\v{s}} Brim, and Thomas Henzinger. 2023.
\newblock Repository of logically consistent real-world boolean network models.
\newblock \emph{bioRxiv}, pages 2023--06.

\bibitem[{Radford et~al.(2021)Radford, Kim, Hallacy, Ramesh, Goh, Agarwal, Sastry, Askell, Mishkin, Clark et~al.}]{radford2021learning}
Alec Radford, Jong~Wook Kim, Chris Hallacy, Aditya Ramesh, Gabriel Goh, Sandhini Agarwal, Girish Sastry, Amanda Askell, Pamela Mishkin, Jack Clark, and 1 others. 2021.
\newblock Learning transferable visual models from natural language supervision.
\newblock In \emph{International conference on machine learning}, pages 8748--8763. PmLR.

\bibitem[{Reddy and Shojaee(2025)}]{reddy2025towards}
Chandan~K Reddy and Parshin Shojaee. 2025.
\newblock Towards scientific discovery with generative ai: Progress, opportunities, and challenges.
\newblock In \emph{Proceedings of the AAAI Conference on Artificial Intelligence}, volume~39, pages 28601--28609.

\bibitem[{Runge et~al.(2022)Runge, Gillies, Strobl, and Palachy-Affek}]{runge2022tigramite}
Jakob Runge, Ewen Gillies, Eric~V Strobl, and Shay Palachy-Affek. 2022.
\newblock Tigramite--causal inference and causal discovery for time series datasets.
\newblock Available at \url{https://github.com/jakobrunge/tigramite?tab=readme-ov-file}.

\bibitem[{Runge et~al.(2019)Runge, Nowack, Kretschmer, Flaxman, and Sejdinovic}]{runge2019detecting}
Jakob Runge, Peer Nowack, Marlene Kretschmer, Seth Flaxman, and Dino Sejdinovic. 2019.
\newblock Detecting and quantifying causal associations in large nonlinear time series datasets.
\newblock \emph{Science advances}, 5(11):eaau4996.

\bibitem[{Shojaee et~al.(2023)Shojaee, Meidani, Barati~Farimani, and Reddy}]{shojaee2023transformer}
Parshin Shojaee, Kazem Meidani, Amir Barati~Farimani, and Chandan Reddy. 2023.
\newblock Transformer-based planning for symbolic regression.
\newblock \emph{Advances in Neural Information Processing Systems}, 36:45907--45919.

\bibitem[{Shojaee et~al.(2024)Shojaee, Meidani, Gupta, Farimani, and Reddy}]{shojaee2024llm}
Parshin Shojaee, Kazem Meidani, Shashank Gupta, Amir~Barati Farimani, and Chandan~K Reddy. 2024.
\newblock Llm-sr: Scientific equation discovery via programming with large language models.
\newblock \emph{arXiv preprint arXiv:2404.18400}.

\bibitem[{Snell et~al.(2024)Snell, Lee, Xu, and Kumar}]{snell2024scaling}
Charlie Snell, Jaehoon Lee, Kelvin Xu, and Aviral Kumar. 2024.
\newblock Scaling llm test-time compute optimally can be more effective than scaling model parameters.
\newblock \emph{arXiv preprint arXiv:2408.03314}.

\bibitem[{Team(2024)}]{qwen2.5}
Qwen Team. 2024.
\newblock \href {https://qwenlm.github.io/blog/qwen2.5/} {Qwen2.5: A party of foundation models}.

\bibitem[{Team(2025)}]{qwen3technicalreport}
Qwen Team. 2025.
\newblock \href {https://arxiv.org/abs/2505.09388} {Qwen3 technical report}.
\newblock \emph{Preprint}, arXiv:2505.09388.

\bibitem[{Wang and Chen(2023)}]{wang2023review}
Jianxun Wang and Yixiang Chen. 2023.
\newblock A review on code generation with llms: Application and evaluation.
\newblock In \emph{2023 IEEE International Conference on Medical Artificial Intelligence (MedAI)}, pages 284--289. IEEE.

\bibitem[{Wang et~al.(2023{\natexlab{a}})Wang, Lauriola, and Moschitti}]{wang2023accurate}
Liang Wang, Ivano Lauriola, and Alessandro Moschitti. 2023{\natexlab{a}}.
\newblock Accurate training of web-based question answering systems with feedback from ranked users.
\newblock In \emph{Proceedings of the 61st Annual Meeting of the Association for Computational Linguistics (Volume 5: Industry Track)}, pages 660--667.

\bibitem[{Wang et~al.(2025)Wang, Wang, Li, Zhang, and Cheng}]{wang2025drsr}
Runxiang Wang, Boxiao Wang, Kai Li, Yifan Zhang, and Jian Cheng. 2025.
\newblock Drsr: Llm based scientific equation discovery with dual reasoning from data and experience.
\newblock \emph{arXiv preprint arXiv:2506.04282}.

\bibitem[{Wang et~al.(2023{\natexlab{b}})Wang, Hu, Lu, Zhu, Zhang, Subramaniam, Loomba, Zhang, Sun, and Wang}]{wang2023scibench}
Xiaoxuan Wang, Ziniu Hu, Pan Lu, Yanqiao Zhu, Jieyu Zhang, Satyen Subramaniam, Arjun~R Loomba, Shichang Zhang, Yizhou Sun, and Wei Wang. 2023{\natexlab{b}}.
\newblock Scibench: Evaluating college-level scientific problem-solving abilities of large language models.
\newblock \emph{arXiv preprint arXiv:2307.10635}.

\bibitem[{Wang et~al.(2024)Wang, Chen, Han, Lin, Zhao, Liu, Zhai, Yuan, You, and Yang}]{wang2024exploring}
Yiqi Wang, Wentao Chen, Xiaotian Han, Xudong Lin, Haiteng Zhao, Yongfei Liu, Bohan Zhai, Jianbo Yuan, Quanzeng You, and Hongxia Yang. 2024.
\newblock Exploring the reasoning abilities of multimodal large language models (mllms): A comprehensive survey on emerging trends in multimodal reasoning.
\newblock \emph{arXiv preprint arXiv:2401.06805}.

\bibitem[{Xie et~al.(2025)Xie, Cui, Zhang, Lu, Shu, Nahab, Hu, and Yang}]{xie2025kerap}
Yuzhang Xie, Hejie Cui, Ziyang Zhang, Jiaying Lu, Kai Shu, Fadi Nahab, Xiao Hu, and Carl Yang. 2025.
\newblock Kerap: A knowledge-enhanced reasoning approach for accurate zero-shot diagnosis prediction using multi-agent llms.
\newblock \emph{arXiv preprint arXiv:2507.02773}.

\bibitem[{Xie et~al.(2024)Xie, Li, He, Xu, Wen, Zhang, Chen, Shi, and Pei}]{xie2024chatts}
Zhe Xie, Zeyan Li, Xiao He, Longlong Xu, Xidao Wen, Tieying Zhang, Jianjun Chen, Rui Shi, and Dan Pei. 2024.
\newblock Chatts: Aligning time series with llms via synthetic data for enhanced understanding and reasoning.
\newblock \emph{arXiv preprint arXiv:2412.03104}.

\bibitem[{Yu et~al.(2011)Yu, Lloyd, Nickerson, Cooling, Miller, Garny, Terkildsen, Lawson, Britten, Hunter et~al.}]{yu2011physiome}
Tommy Yu, Catherine~M Lloyd, David~P Nickerson, Michael~T Cooling, Andrew~K Miller, Alan Garny, Jonna~R Terkildsen, James Lawson, Randall~D Britten, Peter~J Hunter, and 1 others. 2011.
\newblock The physiome model repository 2.
\newblock \emph{Bioinformatics}, 27(5):743--744.

\bibitem[{Zhang et~al.(2024)Zhang, Gao, Liu, and Gao}]{zhang2024logicgep}
Dezhen Zhang, Shuhua Gao, Zhi-Ping Liu, and Rui Gao. 2024.
\newblock Logicgep: Boolean networks inference using symbolic regression from time-series transcriptomic profiling data.
\newblock \emph{Briefings in Bioinformatics}, 25(4):bbae286.

\bibitem[{Zhang et~al.(2025)Zhang, Yang, Han, Qin, and Wang}]{zhang2025tempogpt}
Haochuan Zhang, Chunhua Yang, Jie Han, Liyang Qin, and Xiaoli Wang. 2025.
\newblock Tempogpt: Enhancing temporal reasoning via quantizing embedding.
\newblock \emph{arXiv preprint arXiv:2501.07335}.

\end{thebibliography}

\clearpage
\begin{appendices}
\section{Related Work}

\noindent\textbf{Symbolic structure Discovery.}
\label{appendix: related_work}
Discovering symbolic laws from time series data is a central objective in many scientific discovery tasks. One prominent approach is Symbolic Regression, which seeks closed-form expressions that accurately model the observed data. Classical methods such as Genetic Programming (GP) \cite{mundhenk2021symbolic} are powerful but computationally intensive and often sensitive to the choice of operators and fitness functions. Alternatively, sparse optimization techniques like SINDy~\cite{brunton2016discovering} and PySR~\cite{cranmer2024pysr} aim to identify parsimonious models by leveraging sparsity in the function space.
More recently, deep learning-based models have been introduced to enhance efficiency and scalability. Methods such as ODEformer~\cite{d2023odeformer} and TPSR~\cite{shojaee2023transformer} reformulate symbolic discovery as a sequence-to-sequence translation task, mapping time series data to symbolic equations. These approaches generate high-quality expressions and offer improved computational efficiency. However, they typically require large-scale pretraining and often lack the capability for iterative refinement and adaptation across diverse scientific domains.
Beyond symbolic regression, related tasks such as Boolean Network Inference\cite{zhang2024logicgep} and Causal Discovery\cite{hasan2023survey} also aim to extract symbolic structures from time series data. These methods seek to uncover underlying logical or causal relationships, further emphasizing the broader interest in interpretable, symbolic representations of dynamical systems.

\noindent\textbf{LLM Symbolic Reasoning.}
LLMs exhibit strong in-context learning and reasoning capabilities, enabling them to perform tasks such as logical inference~\cite{ahn2024large, wang2023review} and even preliminary forms of temporal reasoning~\cite{fang2024large}.  Recent LLM systems also benefit from combining structured and textual context~\cite{glenn2024blendsql,lei2025mixture,liu2025hypervectors,wang2023accurate}. Prior work on time series reasoning with LLMs, however, has largely emphasized \textbf{linguistic outputs}, for example, generating natural-language summaries of temporal patterns, identifying anomalies, or extracting descriptive insights from sequential data. While valuable, these approaches focus on surface-level interpretation and do not attempt to uncover the \textbf{underlying symbolic structures} that govern dynamic systems. In contrast, our work targets symbolic reasoning tasks where the output is not a narrative but a formal structure: scientific equations, logical rules, or causal graphs. These forms require abstraction, generalization, and alignment with domain context, going beyond function generation. Existing efforts to apply LLMs to symbolic regression have mainly treated them as equation generators~\cite{merler2024context, li2024mllm, shojaee2024llm}, producing functions that may fit data but often lack explanatory power or contextual plausibility. Moreover, the symbolic space considered so far has been restricted almost exclusively to algebraic expressions, leaving other important forms such as Boolean networks~\cite{zhang2024logicgep} and causal relations~\cite{assaad2022survey} largely unexplored.

\section{SymbolBench Dataset}
\label{appendix: dataset_examples}

To rigorously evaluate the reasoning abilities of LLMs on time series-related science discovery, we introduce \textbf{SymbolBench}, a curated benchmark that aims to uncover symbolic structures from time series. The dataset spans diverse domains (e.g., physics, biology) and is structured around three core categories of symbolic structures. 

\noindent\textbf{Coupled Differential Equations (CDEs).}  
CDEs represent dynamic systems, yielding continuous and multivariate time series data. Let $x_i$ denote the $i$-th variable of a multivariate time series, a coupled differential equation can be described as the following:
$ \frac{dx_i}{dt} = f_i(x_1, x_2, ..., x_n), i = 1, ..., n $,
where the symbolic structures \( f_i \) describe dynamic interactions among state variables. The corresponding time series data consists of numerical solutions \( \{x_i(t)\} \) over time, generated from initial conditions and system parameters. This setting reflects real-world dynamical systems in physics and engineering, where the complexity arises from variable interdependence. While the previous benchmark dataset, ODEbench~\cite{d2023odeformer}, contains coupled ODEs from 1 dimension to 4 dimensions, the number of samples is small, and the class of dimensions is heavily imbalanced, with only three 4-dimensional ODEs. In this study, we further enrich ODEbench with more high-dimensional ODEs and provide a balanced dataset with over 156 samples. Each sample is accompanied by the variable descriptions and the domain name if available.

\noindent\textbf{Boolean Networks (BNs).}  
Multivariate time series with discrete values are also seen in the scientific domains. Derived from models used in systems biology, particularly in gene regulatory and signaling networks, Boolean networks represent each variable as a binary node whose state evolves according to logical expressions:
$x_i^{(t+1)} = f_i(x_1^{(t)}, x_2^{(t)}, ..., x_n^{(t)}), x_i \in \{0, 1\},$
where \( f_i \) is a logical function composed of AND, OR, NOT, etc. The time series data is a sequence of binary vectors over discrete time steps, representing the dynamic evolution of the system. This setup emphasizes symbolic logic reasoning, state transitions, and rule discovery from temporal traces of binary states. In this study, we provide a curated subset of 65 Boolean networks from BioDivine~\cite{pastva2023repository}. For each sample, we provide a short description of the domain and the name of each variable.

\noindent\textbf{Structured Causal Models (SCMs).}  
Beyond specifying a specific data-generating process (e.g., via mathematical functions), studying causal dependencies among time series variables is valuable for uncovering interdependencies directly from raw data, akin to the broader task of causal discovery in temporal settings. This approach models systems using \emph{Structural Causal Models} (SCMs), which can be expressed as a directed graph. For each variable \( x_i \), the goal is to identify its parent variables \( x_j \) along with their corresponding time lags \( l \), such that \( x_j \) at time \( t - l \) causally influences \( x_i \) at time \( t \): $x_i \leftarrow \{ (x_j, l) \mid x_j \in X,; l \in [1, M] \},$
where \( X \) denotes the set of all variables and \( M \) is the maximum considered lag. As the number of variables and potential lag intervals increases, the search space for an optimal SCM grows exponentially, making discovery more challenging. In this work, we extract SCMs from the CDEs in our curated dataset, as well as from additional CDEs in the Physiome database involving more than three variables, using functional analysis. Each sample is annotated with its corresponding SCM, resulting in 190 samples.

SymbolBench uses three different datasets to evaluate the ability of LLMs/MLLMs to uncover the symbolic laws from time series data. We include variable description, domain name, and the time series trajectory as additional context. Examples of the dataset are shown in Table~\ref{tab: dataset_examples}.

\begin{sidewaystable*}[htbp]
\centering
\caption{Summary of datasets, ground truths, and variable context with trajectory images.}
\label{tab: dataset_examples}
\begin{tabular}{|>{\raggedright\arraybackslash}p{3cm}|>{\raggedright\arraybackslash}p{3cm}|>{\raggedright\arraybackslash}p{4cm}|>{\raggedright\arraybackslash}p{3cm}|>{\centering\arraybackslash}p{6cm}|}
\hline
\textbf{Dataset Example} & \textbf{GroundTruth} & \textbf{Variable Description} & \textbf{Domain} & \textbf{Trajectory} \\
\hline

\multirow{3}{=}{Coupled Differential Equation} 
& dx1/dt = (c0 + c1 / (1 + c2 * x2)) - c3 * x1 
& x1: ng\_ml 
& \multirow{3}{=}{Endocrine} 
& \multirow{8}{=}{\includegraphics[width=6cm]{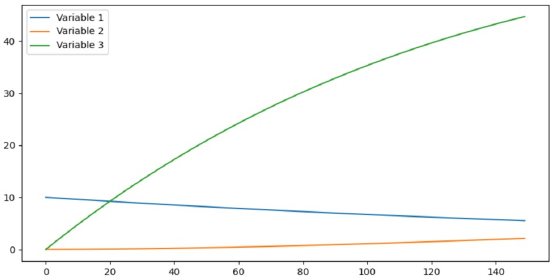}} \\
\cline{2-3}
& dx2/dt = (c8 + c9 * x3) - c10 * x2 
& x2: microg\_dl 
& & \\
\cline{2-3}
& dx3/dt = (c4 + c5 * x1) / (1 + c6 * x2) - c7 * x3 
& x3: pg\_ml 
& & \\
\hline

\multirow{5}{=}{Boolean Network} 
& x1 = ( NOT ( x3 OR x5 ) OR NOT ( x5 OR x3 ) ) 
& x1: v\_Coup\_fti 
& \multirow{5}{=}{Cortical Area Development} 
& \multirow{12}{=}{\includegraphics[width=6cm]{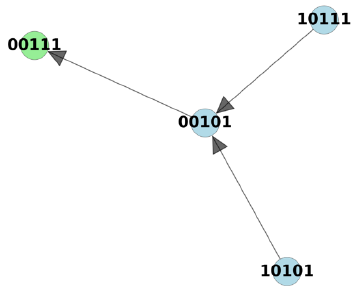}} \\
\cline{2-3}
& x2 = ( x1 AND NOT ( ( x3 OR x5 ) OR x4 ) ) 
& x2: v\_Emx2 
& & \\
\cline{2-3}
& x3 = ( ( x3 AND x5 ) AND NOT x2 ) 
& x3: v\_Fgf8 
& & \\
\cline{2-3}
& x4 = ( x5 AND NOT ( x2 OR x1 ) ) 
& x4: v\_Pax6 
& & \\
\cline{2-3}
& x5 = ( x3 AND NOT x2 ) 
& x5: v\_Sp8 
& & \\
\hline

\multirow{4}{=}{Structured Causal Model} 
& $\sim$ 
& x0: membrane (millivolt) 
& \multirow{4}{=}{Calcium Dynamics} 
& \multirow{6}{=}{\includegraphics[width=4cm]{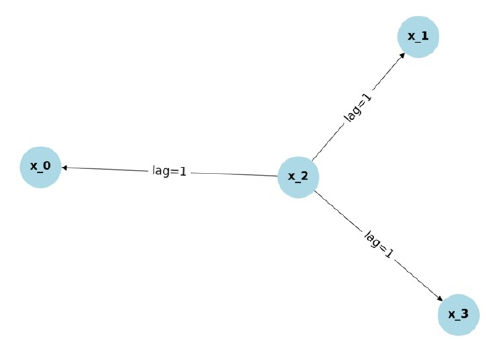}} \\
\cline{2-3}
& $\sim$ 
& \scriptsize x1: rapidly\_activating\_K\_current\_n\_gate (dimensionless) 
& & \\
\cline{2-3}
& $\sim$ 
& x2: Ca\_i in component ionic\_concentrations (micromolar) 
& & \\
\cline{2-3}
& $\sim$ 
& x3: None 
& & \\
\hline

\end{tabular}
\end{sidewaystable*}

\begin{figure}[hbt]
    \centering
    \begin{subfigure}[b]{0.32\textwidth}
        \includegraphics[width=\textwidth]{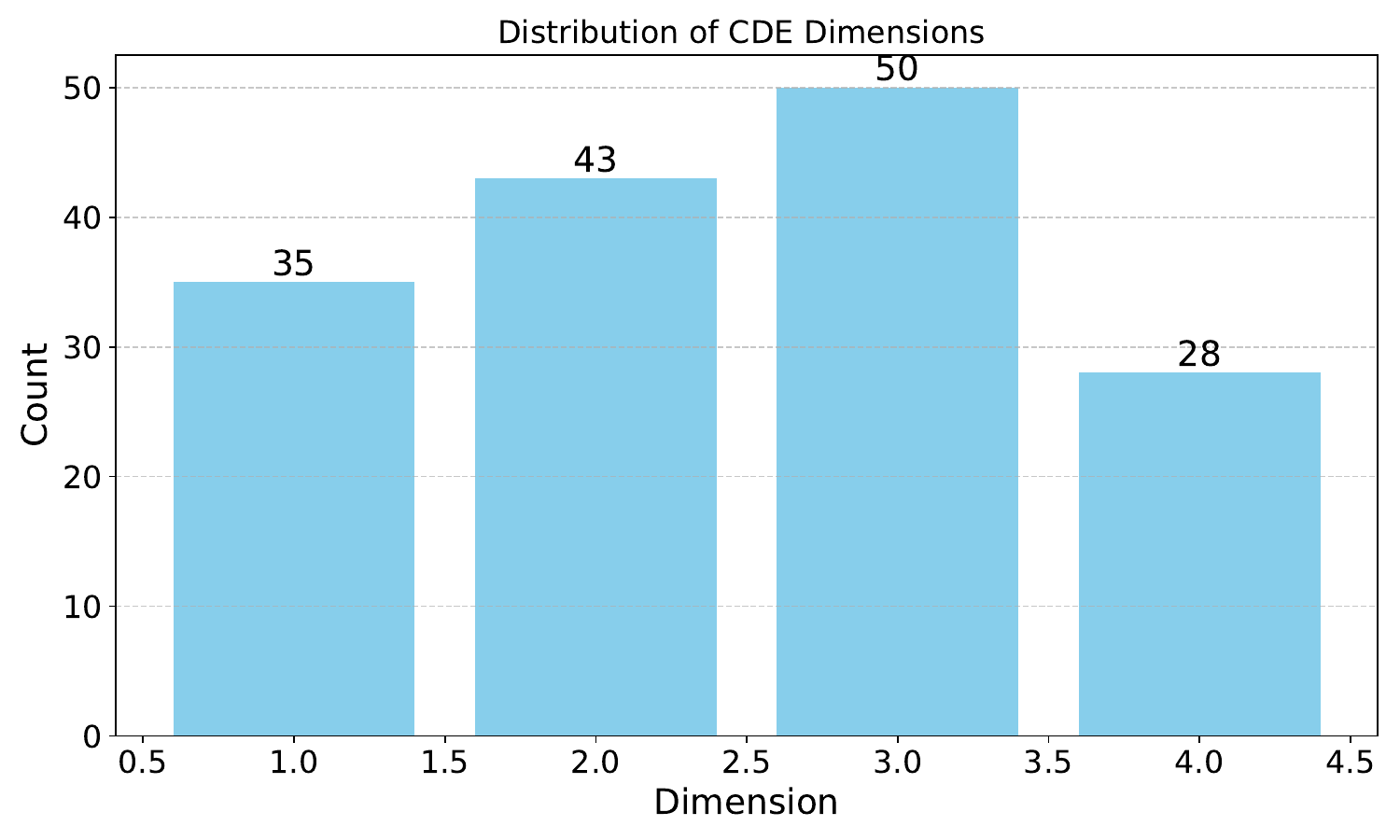}
        \caption{Dimension distribution of Coupled Differential Equations used in SymbolBench.}
        \label{fig: CDEs_distribution}
    \end{subfigure}
    \hfill
    \begin{subfigure}[b]{0.32\textwidth}
        \includegraphics[width=\textwidth]{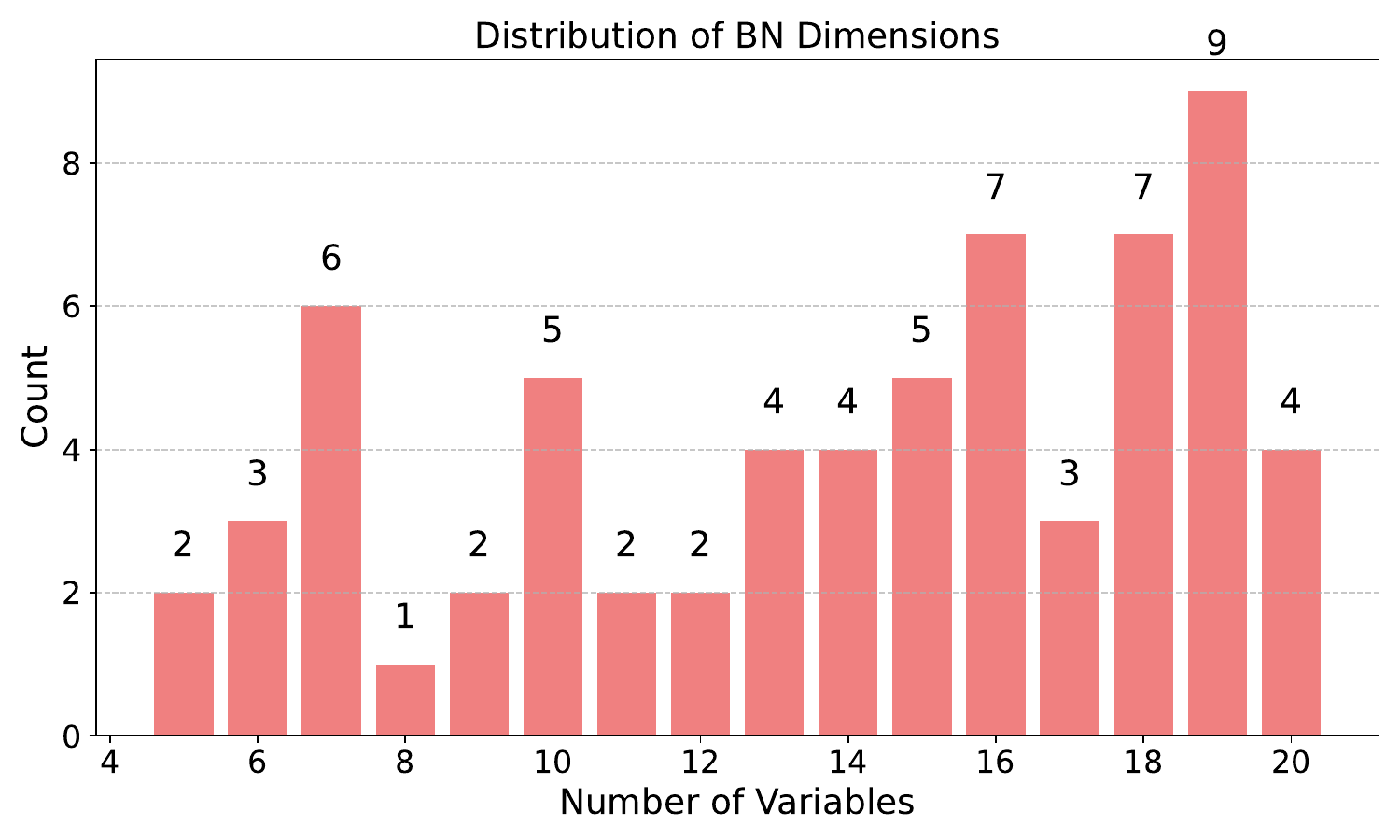}
        \caption{Dimension distribution of Boolean Networks used in SymbolBench.}
        \label{fig: BNs_distribution}
    \end{subfigure}
    \hfill
    \begin{subfigure}[b]{0.32\textwidth}
        \includegraphics[width=\textwidth]{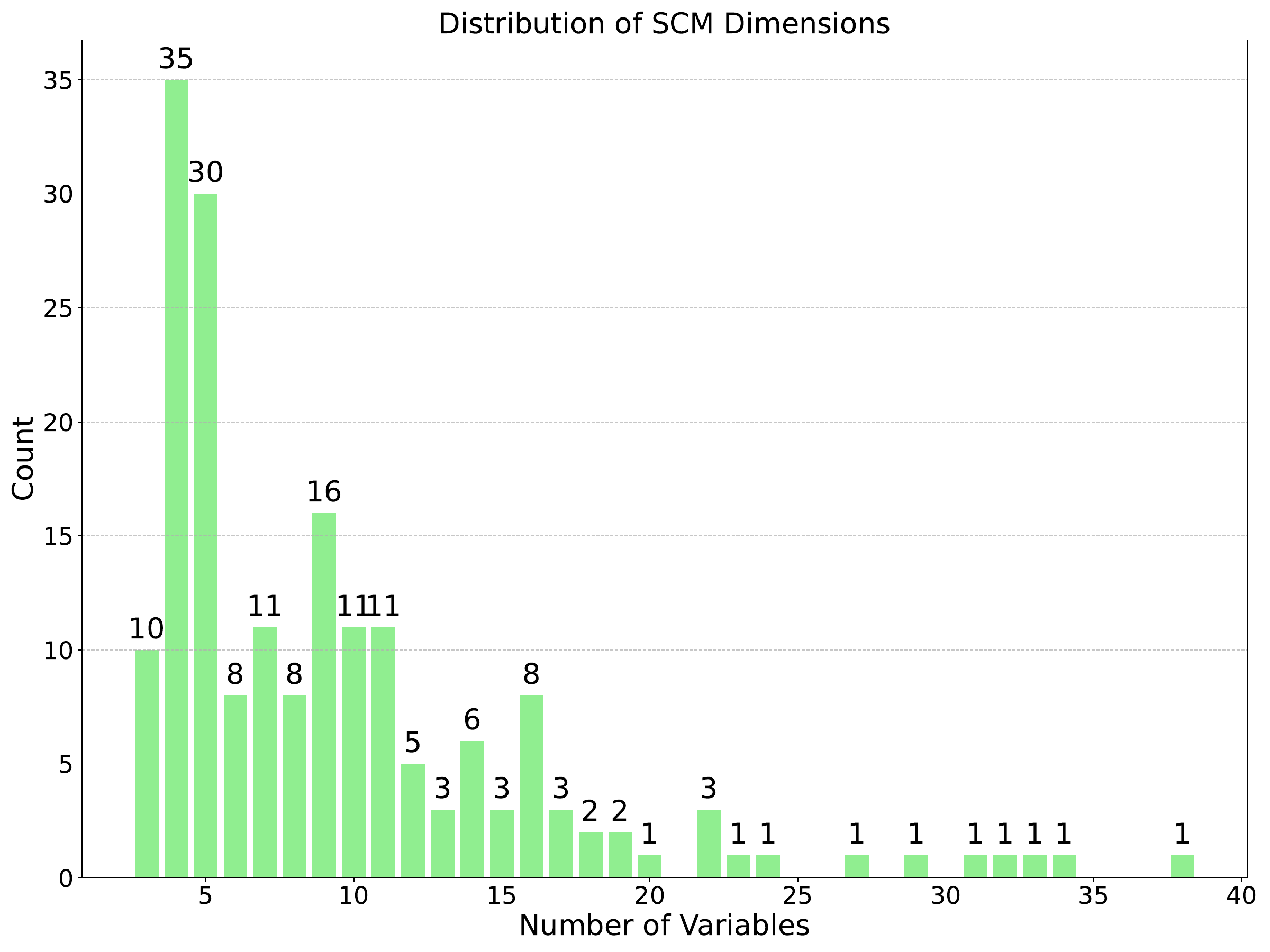}
        \caption{Dimension distribution of Structured Causal Models used in SymbolBench.}
        \label{fig: SCMs_distribution}
    \end{subfigure}
    \label{fig: dim_distributions}
\end{figure}

In addition, we provide a more detailed illustration of the verification process of the three tasks as follows:
\begin{compactenum}[(a)]
    \item \textbf{CDEs:} We use an LLM to generate the skeleton of the continuous‐time dynamical system,
    replacing each unknown coefficient with a placeholder "c". An optimizer then fits these coefficients
    to the training time series. To avoid the high cost of repeatedly solving and differentiating through a
    full ODE solver, we adopt the finite–difference approximation strategy from ODEFormer~\cite{d2023odeformer}.
    Once the functional form $\hat f(\mathbf{x};\boldsymbol{\phi})$ is obtained, where $\phi$ is the fitted coefficients, we generate the final numerical solution using SciPy’s \texttt{scipy.integrate.solve\_ivp}, and compare the trajectory to the ground truth:
    \begin{equation}
    \small
      \texttt{solve\_ivp}\bigl(\hat f(\mathbf{x};\boldsymbol{\phi}),\;\mathbf{x}(t_0),\;t_0,\ldots,t_n, \texttt{method=LSODA}\bigr).
    \end{equation}
    The distribution of sample dims is shown in Figure~\ref{fig: CDEs_distribution}.
    
    \item \textbf{BNs:} For Boolean Networks, the LLM directly outputs a set of logical update rules (e.g.\ $x_i(t+1) = x_j(t)\,\land\,\lnot x_k(t)$). Since there are no continuous parameters to fit, we simply simulate the network from the known initial state $x(t_0)$ and compute the F1 score over all bits and time steps to assess agreement with the true dynamics. The distribution of sample dims is shown in Figure~\ref{fig: BNs_distribution}.
    
    \item \textbf{SCMs:} For Structured Causal Models, the LLM predicts the possible causal relations among all variables, forming a directed graph. Since the predicted SCMs can not directly produce numerical solutions, to quantify how well the predicted SCM explains the data, we compute the sample partial correlation between $x_i$ and each candidate parent in $\mathrm{pa}(x_i)$ (conditioning on the parents), following the protocol of Runge et al.~\cite{runge2019detecting}. The final score for each node is the mean of its absolute partial correlations, and we average over all nodes to obtain the overall SCM score. The distribution of sample dims is shown in Figure~\ref{fig: SCMs_distribution}.
\end{compactenum}

\noindent\textbf{Textual Context.}  
For regular LLMs that only accept textual inputs, all inputs to LLMs are formatted as structured textual prompts. Time series data are serialized into strings, supplemented with contextual metadata such as domain information, variable meanings, and prior scored expressions. This enables language models to reason over both data patterns and contextual priors.

\noindent\textbf{Visual and Temporal Context.}  
Multimodal LLMs have shown promise in handling visual and temporal data. While visual inputs are traditionally used in vision-language tasks~\cite{radford2021learning}, recent research has extended MLLMs to time series domains (e.g., forecasting via visual encodings \cite{li2023time}). Recent models like ChatTS \cite{xie2024chatts} and TempoGPT \cite{zhang2025tempogpt} enable joint reasoning over temporal and textual modalities. In SymbolBench, we explore the use of MLLMs to incorporate both visual time series plots and encoded temporal embeddings.

\section{LLMs for Benchmarking}
We evaluate six representative LLMs, chosen for their diversity in size, training data, and specialization. For all models, we set the temperature=1, top-p=0.9, top-k=60, and the maximum tokens generated to be 4098 to explore diverse and creative symbolic structures. 

\begin{compactenum}[(a)]
    \item \textbf{Qwen Series~\cite{qwen2.5, qwen3technicalreport}:} Qwen2.5-14B~\cite{qwen2.5} is a 14.7-billion-parameter causal Transformer (13.1 B non-embedding) built on RoPE, SwiGLU, RMSNorm, and QKV-bias that extends context support to 128 K tokens (with generation up to 8 K) and delivers significantly richer knowledge, advanced coding, and mathematical reasoning (via domain-expert submodels), robust instruction following, long-form text, and structured output (e.g., tables, JSON), and out-of-the-box multilingual fluency across 29+ languages. Following Qwen2.5-14B, Qwen3~\cite{qwen3technicalreport} series is the newest generation in the Qwen family, combining dense and Mixture-of-Experts architectures to deliver seamless mode-switching—“thinking” for deep logical reasoning, math, and coding, and “non-thinking” for fast, general dialogue.
    
    \item \textbf{Llama 3.2-3B~\cite{dubey2024llama}:} Llama 3.2 is a family of multilingual 1 B and 3 B–parameter pretrained and instruction-tuned text-in/text-out models, with its instruction-tuned versions specially optimized for dialogue, agentic retrieval, and summarization across dozens of languages—consistently outperforming many open-source and proprietary chat models on standard industry benchmarks.
    
    \item \textbf{Mathstral-7B~\cite{jiang2023mistral7b}:} Mathstral is a 7 billion-parameter LLM released by Mistral AI as a tribute to Archimedes’ 2311th anniversary, built on Mistral 7B with a 32K token context window and fine-tuned for advanced multi-step mathematical and scientific reasoning. Developed in collaboration with Project Numina, it achieves state-of-the-art performance for its size.
    
    \item \textbf{ChatTS-14B~\cite{xie2024chatts}:} ChatTS-14B is a multimodal LLM explicitly designed around time series as its core modality, offering native support for multivariate sequences of varying lengths and dimensions, preserving raw numerical fidelity for precise statistical queries, and enabling interactive, conversational exploration and reasoning over time-series data—while also integrating seamlessly into existing LLM workflows (including vLLM) with provided code, datasets, and models.
    
    \item \textbf{GPT-4o-mini~\cite{achiam2023gpt}:} GPT-4o-mini is a compact multimodal reasoning model released by OpenAI in July 2024, delivering GPT-4–level performance while costing over 60\% less than GPT-3.5 Turbo; it supports text and vision inputs, advanced function calling, and extended long-context understanding.
    
    \item \textbf{o4-mini~\cite{achiam2023gpt}:} o4-mini is OpenAI’s latest release of a reasoning-focused GPT variant that replaces o3-mini, offering both text and image processing, “whiteboard” chain-of-thought reasoning, seamless tool integration, and a high-accuracy paid-tier option—all accessible via ChatGPT and the Completions API for domain-critical decision-making tasks.
\end{compactenum}

\section{Baseline Implementation}
\begin{compactenum}[(a)]
    \item \textbf{CDEs:} We follow ODEFormer’s baseline implementation and hyperparameter tuning~\cite{d2023odeformer}, using PySR with finite‐difference approximations for skeleton search, and default greedy top-\(k\) generation for ODEFormer.
    
    \item \textbf{BNs:} We reimplement LogicGep’s Boolean‐network inference using the same Geppy genetic‐programming framework~\cite{gao2022LearningAsynchronousBoolean}, but omit the continuous‐to‐binary discretization and MLP‐based constraint stages, since our training traces are already binary.
    
    \item \textbf{SCMs:} All SCM baselines are based on Tigramite~\cite{runge2022tigramite}, with the maximum time‐lag set to 1 for fair comparison.
\end{compactenum}







\section{Full Results}
\label{appendix: full_bn_scm}
We present full results across all dimensions in Table~\ref{tab: full-symbolic}, and the performance of more LLMs on BN and SCM discovery in Table~\ref{tab: full_BN_} and Table~\ref{tab: full_SCM_}.

\begin{table*}[ht]
\centering
\caption{Symbolic regression results for CDEs across 4 dimensions. We use percentage for $SR^2$ and $ACC_{0.9}$. \textcolor{orange}{Orange},\textcolor{yellow}{Yellow}, and \textcolor{cyan}{Cyan} mark the first, second, and third place, respectively. 
Statistically significant improvements over the best baseline (paired t-test, $p$ < 0.05) are \underline{underlined}.}
\vskip -0.8em
\label{tab: full-symbolic}
\renewcommand{\arraystretch}{1}
\setlength{\tabcolsep}{0.5pt}
\begin{threeparttable}
\fontsize{7.5}{9}\selectfont
\begin{tabular}{
c l l
*{3}{S[table-format=3.2]}
*{4}{S[table-format=3.2]}
*{2}{S[table-format=3.2]}
*{2}{S[table-format=3.2]}
*{2}{S[table-format=3.2]}
*{2}{S[table-format=3.2]}
*{2}{S[table-format=3.2]}
}
\toprule
\multirow{2}{*}{Dim} & \multirow{2}{*}{} & \multirow{2}{*}{Metric}
  & \multicolumn{3}{c}{Baselines}
  & \multicolumn{4}{c}{Qwen2.5}
  & \multicolumn{2}{c}{Llama3.2}
  & \multicolumn{2}{c}{Mathstral}
  & \multicolumn{2}{c}{4o-text}
  & \multicolumn{2}{c}{4o-image}
  & \multicolumn{2}{c}{ChatTS} \\
\cmidrule(lr){4-6}\cmidrule(lr){7-10}\cmidrule(lr){11-12}\cmidrule(lr){13-14}
\cmidrule(lr){15-16}\cmidrule(lr){17-18}\cmidrule(lr){19-20}
 &  & 
 & {Pro.} & {PySR} & {ODE.}
 & {Na\"ive} & {Base} & {Ctx} & {CoT}
 & {Ctx} & {CoT}
 & {Ctx} & {CoT}
 & {Ctx} & {CoT}
 & {Ctx} & {CoT}
 & {Ctx} & {CoT} \\
\midrule
\multirow{6}{*}{\rotatebox[origin=c]{90}{Dim=1}}
  & \multicolumn{2}{l}{Complexity $\downarrow$}                 & 5.06 & 2.91 & 4.68 & 4.17 & 5.14 & 5.34 & 3.91 & \bestone{1.89} & \besttwo{1.97} & 2.63 & 2.69 & 2.31 & 2.36 & 2.37 & \bestthree{2.09} & 3.29 & 3.00 \\
  & \multicolumn{2}{l}{Symbolic Prox. $\downarrow$}             & 4.86 & \bestone{3.54} & 5.18 & 5.11 & 5.37 & 5.57 & 4.89 & 4.03 & 4.14 & \bestthree{3.86} & 4.09 & 3.89 & \besttwo{3.79} & 3.97 & 4.06 & 4.83 & 4.37 \\
  & \multirow{2}{*}{ID}  & $SR^2$                     & 95.50 & 96.90 & 83.40 & 93.30 &  \underline{97.20} & \bestone{\underline{99.00}} & \besttwo{\underline{97.40}} & 95.10 & 94.90 & 96.20 & 96.20 & 92.80 & 95.40 & 95.70 & 95.30 & 95.10 & \bestthree{97.00} \\
  &                       & $ACC_{0.9}$                   & 91.40 & \besttwo{97.10} & 73.50 & 91.40 & \besttwo{97.10} & \bestone{\underline{100.00}} & \besttwo{97.10} & 91.40 & 91.40 & \besttwo{97.10} & \besttwo{97.10} & 94.30 & \bestthree{97.00} & \besttwo{97.10} & \besttwo{97.10} & 91.40 & \besttwo{97.10} \\
  & \multirow{2}{*}{OOD} & $SR^2$                     & \besttwo{74.10} & \bestone{88.20} & 54.10 & 60.90 & 74.00 & 61.80 & 61.80 & 69.90 & 64.60 & 71.00 & 66.90 & 70.80 & 66.20 & \bestthree{71.50} & 62.90 & 63.80 & 70.30 \\
  &                       & $ACC_{0.9}$                   & \besttwo{65.70} & \bestone{85.70} & 47.10 & 45.70 & \bestthree{63.60} & 52.90 & 51.40 & 51.40 & 48.60 & 62.90 & 60.00 & 54.30 & 51.50 & 54.30 & 45.70 & 51.40 & 54.30 \\
\cmidrule(lr){2-20}
 & \multicolumn{2}{l}{Avg. ID Rank $\downarrow$}  & 10.5 & 3.5 & 17.0 & 13.5 & 2.5	& 1.0	& 2.0	& 12.0 & 13.0 & 4.0 & 4.0 & 13.5 & 10.0 & 5.0 & 6.5	& 12.0 & 3.0 \\
 & \multicolumn{2}{l}{Avg. OOD Rank $\downarrow$} & 2.0 & 1.0 & 16.0 & 16.0 & 3.0 & 11.5 & 12.5 & 9.5 & 12.5 & 4.5 & 7 & 6 & 10 & 5 & 14.5 & 11.5 & 6.5 \\
\midrule
\multirow{6}{*}{\rotatebox[origin=c]{90}{Dim=2}}
  & \multicolumn{2}{l}{Complexity $\downarrow$}                 & \bestone{5.71} & 7.00 & 9.05 & 8.76 & 9.75 & 10.00 & 9.07 & 6.93 & 7.07 & 7.93 & 8.35 & 6.91 & 6.64 & \besttwo{6.33} & \besttwo{6.33} & 9.88 & 9.63 \\
  & \multicolumn{2}{l}{Symbolic Prox. $\downarrow$}             & 10.90 & \bestone{8.21} & \bestthree{9.98} & 11.20 & 11.60 & 11.40 & 11.10 & \besttwo{10.00} & 9.86 & 10.30 & 10.40 & 10.40 & \bestthree{9.98} & 10.10 & \besttwo{10.00} & 11.70 & 11.40 \\
  & \multirow{2}{*}{ID}  & $SR^2$                     & 61.20 & 80.40 & 72.50 &  \underline{82.40} &  \underline{87.90} &  \underline{87.40} & 88.00 & 82.20 &  \underline{80.90} & \besttwo{\underline{88.60}} & \bestthree{\underline{88.40}} & 78.50 & 77.40 & 78.40 & 82.40 &  $\underline{84.80}$ & \bestone{\underline{90.90}} \\
  &                       & $ACC_{0.9}$                   & 42.90 & 76.90 & 47.60 & 71.10 & 77.30 & \besttwo{\underline{80.00}} & 76.70 & 65.10 & 65.10 & \bestthree{79.10} & \bestthree{79.10} & 62.20 & 59.50 & 62.80 & 62.80 & 76.70 & \bestone{83.70} \\
  & \multirow{2}{*}{OOD} & $SR^2$                     & 18.80 & \bestone{54.00} & 34.60 & 44.90 & \besttwo{52.90} & 45.30 & \bestthree{49.70} & 40.60 & 47.30 & 46.10 & 46.80 & 42.70 & 40.20 & 38.10 & 45.60 & 49.50 & 44.70 \\
  &                       & $ACC_{0.9}$                   & 12.50 & \bestthree{42.10} & 18.60 & 31.00 & \bestone{\underline{44.40}} & 34.10 & 35.90 & 32.60 & 34.90 & 34.90 & 33.30 & 29.50 & 22.50 & 23.80 & 29.30 & \besttwo{43.60} & 33.30 \\
\cmidrule(lr){2-20}
 & \multicolumn{2}{l}{Avg. ID Rank $\downarrow$}  & 17.0 & 9.0 & 16.0 & 8.5 & 5.0 & 4.0 & 5.5 & 10.0 & 10.5 & 2.5 & 3.0 & 13.5 & 15.0 & 13.0 & 10.0 & 7.0 & 1.0 \\
 & \multicolumn{2}{l}{Avg. OOD Rank $\downarrow$} & 17.0 & 2.0 & 16.0 & 10.5 & 1.5 & 8.0 & 3.5 & 11.5 & 5.0 & 6.0 & 7.0 & 12.0 & 14.5 & 14.5 & 10.5 & 3.0 & 9.5 \\
\midrule
\multirow{6}{*}{\rotatebox[origin=c]{90}{Dim=3}}
  & \multicolumn{2}{l}{Complexity $\downarrow$}                 & \bestone{7.87} & \besttwo{8.82} & 13.90 & 16.60 & 17.40 & 18.30 & 15.50 & 15.70 & 17.30 & 16.30 & 15.40 & 12.60 & 12.30 & 12.00 & \bestthree{11.50} & 17.00 & 17.20 \\
  & \multicolumn{2}{l}{Symbolic Prox. $\downarrow$}             & 30.30 & \bestone{25.70} & 30.50 & 29.80 & 30.10 & 29.70 & \bestthree{28.60} & 28.80 & 30.00 & 29.50 & 29.00 & 29.50 & \besttwo{27.40} & 29.20 & 29.20 & 29.40 & 29.60 \\
  & \multirow{2}{*}{ID}  & $SR^2$                     & 20.90 & 67.10 & 36.30 & \bestthree{71.30} & \besttwo{74.30} & \bestone{\underline{75.30}} & 70.90 & 62.10 & 57.40 & 65.20 & 64.60 &  \underline{70.00} & 63.70 & 65.30 & 66.90 & 70.80 & 68.40 \\
  &                       & $ACC_{0.9}$                   & 10.90 & 60.50 & 18.40 & \bestthree{61.20} & \besttwo{63.30} & \bestone{\underline{80.00}} & 53.10 & 49.00 & 43.50 & 57.10 & 57.10 & 56.30 & 50.00 & 49.00 & 53.10 & 59.20 & 57.10 \\
  & \multirow{2}{*}{OOD} & $SR^2$                     & 8.60 & \bestone{49.90} & 30.00 & \besttwo{47.20} & 43.20 & 44.30 & 40.90 & 43.10 & 35.90 & 38.70 & \bestthree{45.00} & 44.70 & 39.40 & 39.20 & 37.30 & 40.20 & 42.40 \\
  &                       & $ACC_{0.9}$                   & 4.50 & 36.10 & 15.20 & \besttwo{39.50} & \bestthree{38.60} & 36.60 & 33.30 & 34.00 & 28.60 & 33.30 & \bestone{40.90} & 38.30 & 31.70 & 31.10 & 29.20 & 32.60 & 33.30 \\
\cmidrule(lr){2-20}
 & \multicolumn{2}{l}{Avg. ID Rank $\downarrow$}  & 17.0 & 6.0 & 16.0 & 3.0 & 2.0 & 1.0 & 7.0 & 13.5 & 15.0 & 8.5 & 9.0 & 7.5 & 12.5 & 11.5 & 9.5 & 5.0 & 6.5 \\
 & \multicolumn{2}{l}{Avg. OOD Rank $\downarrow$} & 17.0 & 3.5 & 16.0 & 2.0 & 4.5 & 5.0 & 8.5 & 7.0 & 15.0 & 10.5 & 2.0 & 4.0 & 11.5 & 12.5 & 14.0 & 10.5 & 8.0 \\
\midrule
\multirow{6}{*}{\rotatebox[origin=c]{90}{Dim=4}}
  & \multicolumn{2}{l}{Complexity $\downarrow$}                 & \bestone{9.70} & \besttwo{10.60} & 14.40 & 22.90 & 23.20 & 23.10 & 18.50 & 20.60 & 19.60 & 17.90 & 18.70 & 14.50 & 13.90 & 13.90 & \bestthree{13.60} & 22.60 & 22.10 \\
  & \multicolumn{2}{l}{Symbolic Prox. $\downarrow$}             & 32.70 & \bestone{30.30} & 36.40 & 35.20 & 35.40 & 34.10 & 32.90 & 34.70 & 32.70 & 32.50 & 32.40 & 34.80 & 32.90 & \bestthree{32.20} & \besttwo{31.90} & 38.00 & 36.30 \\
  & \multirow{2}{*}{ID}  & $SR^2$                     & 11.50 & 74.60 & 34.80 & \bestthree{\underline{92.70}} & \bestone{\underline{93.60}} & \besttwo{\underline{93.30}} &  \underline{91.00} & 49.40 & 61.10 & 78.00 & 78.70 & 88.60 & 86.90 & 86.10 & 89.10 & 78.70 & 80.00 \\
  &                       & $ACC_{0.9}$                   & 8.70 & 65.20 & 29.60 &  \underline{87.90} & \bestone{\underline{94.10}} & \besttwo{\underline{93.90}} & \bestthree{\underline{89.30}} & 40.70 & 52.00 & 74.10 & 75.00 & 78.80 & 70.00 & 71.40 & 85.70 & 74.10 & 80.80 \\
  & \multirow{2}{*}{OOD} & $SR^2$                     & 7.00 & 23.80 & 11.40 & 29.20 & 19.10 & 19.40 & {\besttwo{\underline{34.60}}} & 28.30 & 21.30 & \bestthree{\underline{30.70}} & \bestone{\underline{36.90}} &  \underline{26.90} &  \underline{27.60} & 30.40 & 30.50 & 25.00 & 24.50 \\
  &                       & $ACC_{0.9}$                   & 0.00 & 21.70 & 7.10 & 19.40 & 15.60 & 12.50 & {\bestone{\underline{33.30}}} & 12.00 & 13.60 & \bestthree{24.00} & \besttwo{30.80} & 19.40 & 20.70 & 15.40 & 21.40 & 21.70 & 17.40 \\
\cmidrule(lr){2-20}
 & \multicolumn{2}{l}{Avg. ID Rank $\downarrow$}  & 17.0 & 13.0 & 16.0 & 3.5 & 1.0 & 2.0 & 3.5 & 15.0 & 14.0 & 10.5 & 9.0 & 6.5 & 9.5 & 9.5 & 5.0 & 9.5 & 7.5 \\
 & \multicolumn{2}{l}{Avg. OOD Rank $\downarrow$} & 17.0 & 8.0 & 16.0 & 7.0 & 13.0 & 14.0 & 1.5 & 11.0 & 13.0 & 3.0 & 1.5 & 8.5 & 7.5 & 8.5 & 5.0 & 7.0 & 10.5 \\
\bottomrule
\end{tabular}
\end{threeparttable}
\end{table*}

\begin{table*}[t]
\centering
\caption{Comparison of Boolean network inference across ID and OOD settings.}
\label{tab: full_BN_}
\setlength{\tabcolsep}{2pt}
\renewcommand{\arraystretch}{1.1}
\resizebox{\textwidth}{!}{%
\begin{tabular}{c c ccccccc ccccccc c c}
\toprule
\multirow{2}{*}{\textbf{Model}} & \multirow{2}{*}{\textbf{Setting}}
& \multicolumn{7}{c}{\textbf{ID}}
& \multicolumn{7}{c}{\textbf{OOD}}
& \multirow{2}{*}{\textbf{Symb. Prox.} $\downarrow$}
& \multirow{2}{*}{\textbf{Comp.} $\downarrow$} \\
\cmidrule(lr){3-9}\cmidrule(lr){10-16}
& & \textbf{Prec.} & \textbf{Rec.} & \textbf{Acc.} & \textbf{B.I.} & $\mathbf{ACC_{0.5}}$ & $\mathbf{ACC_{0.7}}$ & $\mathbf{ACC_{0.8}}$
  & \textbf{Prec.} & \textbf{Rec.} & \textbf{Acc.} & \textbf{B.I.} & $\mathbf{ACC_{0.5}}$ & $\mathbf{ACC_{0.7}}$ & $\mathbf{ACC_{0.8}}$
  & & \\
\midrule
\multirow{1}{*}{LogicGep}
& $\sim$         & \bestone{93.6} & \bestone{92.7} & \bestone{95.2} & \bestone{88.7} & \bestone{98.5} & \bestone{98.5} & \bestone{96.9}
                 & \bestone{84.7} & \bestone{86.5} & \bestone{89.5} & \bestone{76.5} & \bestone{98.5} & \bestone{86.2} & \bestone{76.9}
                 & 12.39 & \bestone{12.39} \\
\midrule
\multirow{4}{*}{Qwen2.5-14B}
& Na\"ive        & \besttwo{58.7} & 71.7 & \bestthree{66.5} & \bestthree{29.5} & 86.2 & \bestthree{33.8} & \bestthree{6.2}
                 & \besttwo{56.4} & 71.3 & \bestthree{64.5} & 26.6 & 80.0 & \bestthree{24.6} & \bestthree{7.7}
                 & 12.87 & \bestthree{14.84} \\
& Base           & 54.7 & 73.3 & 63.5 & 27.3 & 80.0 & 20.0 & 3.1
                 & 53.8 & 72.8 & 62.5 & 25.9 & 83.1 & 16.9 & 3.1
                 & 12.33 & 16.73 \\
& Context        & 57.8 & \besttwo{77.2} & \besttwo{67.1} & 29.1 & \bestthree{87.7} & \besttwo{38.5} & \besttwo{9.2}
                 & 56.1 & \besttwo{77.2} & \besttwo{65.5} & \bestthree{27.0} & \besttwo{87.7} & \besttwo{30.8} & \besttwo{10.8}
                 & 12.39 & 16.86 \\
& CoT            & \bestthree{58.3} & 73.9 & 65.4 & \besttwo{30.3} & \besttwo{92.3} & 27.7 & 4.6
                 & \bestthree{56.2} & \bestthree{73.9} & 63.2 & \besttwo{27.9} & \bestthree{84.6} & \bestthree{24.6} & 3.1
                 & \besttwo{11.96} & \besttwo{14.79} \\
\midrule
\multirow{2}{*}{Llama3.2-3B}
& Context        & 51.6 & \bestthree{74.8} & 62.4 & 24.5 & 60.0 & 18.5 & 1.5
                 & 47.4 & 71.1 & 58.2 & 16.9 & 52.3 & 9.2 & 1.5
                 & 14.12 & 17.67 \\
& CoT            & 52.0 & 73.1 & 62.3 & 23.0 & 61.5 & 20.0 & 1.5
                 & 49.9 & 67.8 & 58.3 & 15.7 & 56.9 & 13.8 & 1.5
                 & 14.30 & 19.59 \\
\midrule
\multirow{2}{*}{Mathstral-7B}
& Context        & 49.5 & 73.3 & 60.3 & 19.2 & 50.8 & 15.4 & 1.5
                 & 48.0 & 72.2 & 58.0 & 16.4 & 53.8 & 10.8 & 1.5
                 & 12.74 & 20.14 \\
& CoT            & 51.3 & 74.4 & 62.6 & 23.9 & 61.5 & 21.5 & 3.1
                 & 48.3 & 71.7 & 58.3 & 16.9 & 60.0 & 16.9 & 3.1
                 & \bestthree{12.16} & 20.48 \\
\midrule
\multirow{2}{*}{GPT-4o-mini}
& Context        & 51.6 & 58.9 & 62.6 & 21.2 & 47.7 & 15.4 & 0.0
                 & 51.3 & 58.5 & 61.4 & 20.0 & 53.8 & 13.8 & 0.0
                 & 13.24 & 30.07 \\
& CoT            & 52.2 & 58.9 & 63.5 & 23.5 & 53.8 & 4.6 & 0.0
                 & 51.7 & 60.6 & 62.2 & 23.0 & 53.8 & \bestthree{7.7} & 0.0
                 & \bestone{11.20} & 32.42 \\
\bottomrule
\end{tabular}}
\end{table*}

\begin{table*}[t]
\centering
\caption{Performance comparison of causal discovery methods and LLM-based approaches.}
\label{tab: full_SCM_}
\setlength{\tabcolsep}{3pt}
\renewcommand{\arraystretch}{0.8}
\resizebox{0.9\textwidth}{!}{%
\begin{tabular}{c c c c c c c c c c c}
\toprule
\textbf{Model} & \textbf{Setting} & \textbf{F1} & \textbf{Prec.} & \textbf{Recall} & \textbf{FDR$\downarrow$} & $\mathbf{ACC_{0.5}}$ & $\mathbf{ACC_{0.7}}$ & $\mathbf{ACC_{0.8}}$ & \textbf{SHD$\downarrow$} & \textbf{Complx$\downarrow$} \\
\midrule
PCMCI          & $\sim$           & 52.7 & 52.2 & \bestone{63.7} & 46.1 & 55.3 & 23.7 & 7.9  & 95.28 & 96.76 \\
LPCMCI         & $\sim$           & 52.0 & \besttwo{68.6} & 45.9 & \besttwo{29.1} & \bestthree{56.3} & 18.4 & 5.3  & \bestone{25.35} & \bestthree{14.54} \\
j-PCMCI+       & $\sim$           & 46.2 & 58.1 & 43.3 & 39.1 & 45.3 & 13.2 & 7.4  & 49.36 & 38.83 \\
\midrule
\multirow{4}{*}{Qwen2.5-14B}
               & Na\"ive          & 49.7 & 59.4 & 45.0 & 40.6 & 43.7 & 17.9 & 8.4  & 35.54 & 18.93 \\
               & Base             & 50.5 & 61.4 & 45.0 & 38.6 & 47.4 & 18.4 & \bestthree{8.9}  & 34.58 & 18.10 \\
               & Context          & \bestthree{53.4} & \bestthree{63.0} & 48.3 & \bestthree{37.0} & 53.7 & 19.5 & 7.9  & \bestthree{31.69} & 18.61 \\
               & CoT              & 51.3 & 61.9 & 46.3 & 38.1 & 52.6 & \bestthree{20.0} & 8.4  & 39.36 & 21.76 \\
\midrule
\multirow{2}{*}{Llama3.2-3B}
               & Context          & 51.3 & 56.1 & \bestthree{49.0} & 43.9 & 37.4 & 13.7 & \bestthree{8.9}  & 41.89 & 25.43 \\
               & CoT              & 51.8 & 56.4 & \besttwo{49.5} & 43.6 & 48.4 & \besttwo{22.6} & \bestone{11.6} & 44.40 & 25.75 \\
\midrule
\multirow{2}{*}{Mathstral-7B}
               & Context          & 47.2 & 54.8 & 43.1 & 45.2 & 40.5 & 15.8 & 7.9  & 37.70 & 20.48 \\
               & CoT              & 49.8 & 58.0 & 45.4 & 42.0 & 44.7 & 14.2 & 5.8  & 34.30 & 19.38 \\
\midrule
\multirow{2}{*}{GPT-4o-mini}
               & Context          & 36.9 & 59.9 & 27.6 & 40.1 & 24.7 & 5.3  & 1.6  & 44.29 & \besttwo{13.36} \\
               & CoT              & 37.1 & 57.5 & 28.5 & 42.5 & 24.2 & 3.2  & 2.6  & 38.90 & \bestone{13.08} \\
\midrule
\multirow{2}{*}{ChatTS-14B}
               & Context          & \besttwo{54.1} & \bestone{72.3} & 46.2 & \bestone{27.7} & \besttwo{58.9} & \besttwo{22.6} & \bestone{11.6} & 32.48 & 15.07 \\
               & CoT              & \bestone{54.4} & \bestone{72.3} & 46.6 & \bestone{27.7} & \bestone{61.1} & \bestone{25.3} & \besttwo{10.5} & \besttwo{31.10} & 14.70 \\
\bottomrule
\end{tabular}}
\vskip -0.8em
\end{table*}

\section{Error Analysis for Boolean Network Inference}
\label{appendix: BN_error}
Summarizing the performance on Boolean Network tasks, we categorize the errors from the following perspectives.

\subsection{Limited Observations}
Limited observation brings uncertainty in predicting the accurate form of the symbolic structure.

\noindent\textbf{Example:}

\noindent\textbf{Complete Ground Truth Regulations:}
\begin{align}
\text{x1} &= (\neg (x3 \vee x5) \vee \neg (x5 \vee x3)) \\
\text{x2} &= (x1 \wedge \neg ((x3 \vee x5) \vee x4)) \\
\text{x3} &= ((x3 \wedge x5) \wedge \neg x2)
\end{align}

\noindent\textbf{Complete Predicted Regulations:}
\begin{align}
x1 &= (x2 \vee \neg x3) \wedge (x4 \vee \neg x5) \\
x2 &= x2 \wedge x3 \wedge x5 \wedge \neg x3 \\
x3 &= x3 \wedge x5
\end{align}

\noindent\textbf{Analysis:}  
This case demonstrates overfitting due to limited observations. The Boolean network has 5 variables with only 11 training transitions. The model achieves ID F1 = 0.571 but drops to OOD F1 = 0.087 (performance drop: 0.484). The predicted regulations show different logical structures that work for training data but fail to capture true Boolean network dynamics.

\subsection{Aligning Symbolic Structure and Numerical Fitness}
The numerical fitness in the OOD setting is largely determined by the symbolic structure accuracy. One difficulty during the iterative search process is that the proximity of symbolic structure is not correlated with the proximity of numerical fitness, which means numerically similar predictions can have a large difference in symbolic structure similarity.

\noindent\textbf{Example:}

\noindent\textbf{Complete Ground Truth Structure:}
{
\small
\begin{align}
\text{x1} &= (((((((\neg x2 \wedge \neg x4) \wedge x7) \vee (((\neg x2 \wedge x4) \wedge \neg \ldots \\
\text{x2} &= (((((\neg x2 \wedge \neg x4) \vee ((\neg x2 \wedge x4) \wedge x1)) \vee ((x2 \wedge \ldots \\
\text{x3} &= ((x2 \wedge \neg x4) \vee ((x2 \wedge x4) \wedge x1)) \\
\text{x4} &= ((\neg x2 \vee (x2 \wedge \neg x3)) \vee (((x2 \wedge x3) \wedge x4) \wedge x5)) \\
\text{x5} &= (((\neg x2 \wedge x4) \vee ((x2 \wedge \neg x3) \wedge x4)) \vee ((((x2 \wedge x \ldots \\
\text{x6} &= ((((\neg x2 \wedge x4) \wedge x5) \wedge \neg x6) \vee ((((x2 \wedge \neg x3) \wedge \ldots \\
\text{x7} &= ((\neg x2 \wedge \neg x4) \vee ((\neg x2 \wedge x4) \wedge \neg x5))
\end{align}
}

\noindent\textbf{Complete Predicted Structure:}
\begin{align}
x1 &= x7 \\
x2 &= x3 \vee x4 \\
x3 &= x2 \wedge x6 \\
x4 &= x6 \vee x7 \\
x5 &= x3 \vee x4 \\
x6 &= x5 \\
x7 &= x7
\end{align}

\noindent\textbf{Analysis:}  
This case exemplifies structure-fitness misalignment where high NED = 17.43 indicates structural dissimilarity to ground truth, yet F1 performance is good (0.671). This demonstrates that functionally equivalent Boolean regulations can have very different symbolic representations. The model found an alternative solution that achieves similar numerical fitness despite having a completely different structural form. This highlights that tree edit distance may not capture functional equivalence in Boolean networks, where multiple valid regulatory patterns can produce equivalent behaviors.

\subsection{Error in Context Alignment}
While we use context to shrink the search space and hope to predict more accurate symbolic forms, a misalignment with the given context can produce the symbolic structure from the wrong domain.

\noindent\textbf{Example:}

\noindent\textbf{Expected Boolean Network Structure:}
\begin{align}
\text{x1} &= (\neg (x3 \vee x5) \vee \neg (x5 \vee x3)) \\
\text{x2} &= (x1 \wedge \neg ((x3 \vee x5) \vee x4)) \\
\text{x3} &= ((x3 \wedge x5) \wedge \neg x2) \\
\text{x4} &= (x5 \wedge \neg (x2 \vee x1)) \\
\text{x5} &= (x3 \wedge \neg x2)
\end{align}

\noindent\textbf{Predicted Structure:}
\begin{align}
x1 &= (x2 \vee \neg x3) \wedge (x4 \vee \neg x5) \\
x2 &= x2 \wedge x3 \wedge x5 \wedge \neg x3 \\
x3 &= x3 \wedge x5 \\
x4 &= x3 \wedge x4 \\
x5 &= x4 \wedge x5
\end{align}

\noindent\textbf{Biological Domain Context:} This Boolean network models cortical area development during embryonic brain formation. The variables represent critical developmental genes with established biological functions:
\begin{compactenum}[(a)]
\item x1 (v\_Coup\_fti): Transcription factor controlling cortical area identity
\item x2 (v\_Emx2): Homeobox gene regulating cortical arealization  
\item x3 (v\_Fgf8): Growth factor signaling from anterior neural ridge
\item x4 (v\_Pax6): Master regulator of cortical neurogenesis
\item x5 (v\_Sp8): Transcription factor in frontal cortex specification
\end{compactenum}

\noindent\textbf{Analysis:} 
The biological context constrains valid regulatory logic: Fgf8 and Sp8 typically act as upstream signaling molecules that inhibit posterior genes (Coup\_fti, Emx2) while promoting anterior development. Pax6 functions as a competence factor that enables context-dependent responses to signaling gradients. This case demonstrates severe context misalignment with F1 = 0.087, showing fundamental violations of both computational and biological domain constraints:
\begin{compactenum}[(a)]
    \item \textbf{Logical Contradictions:} The predicted x2 regulation (x2 $\wedge$ x3 $\wedge$ x5 $\wedge$ $\neg$x3) contains impossible Boolean conditions, violating basic computational domain rules where a variable cannot be simultaneously true and false.
    
    \item \textbf{Biological Hierarchy Inversion:} The ground truth implements the established developmental cascade where anterior signals (x3: Fgf8, x5: Sp8) suppress posterior genes (x1: Coup\_fti, x2: Emx2). However, the prediction creates feedback loops where x2 positively regulates x1, contradicting known developmental biology where these genes have antagonistic relationships.
    
    \item \textbf{Signaling Logic Violations:} The ground truth correctly models mutual inhibition between anterior (x3, x5) and posterior (x1, x2) gene programs. The prediction fails to capture this essential developmental principle, instead creating cooperative interactions that would be biologically impossible during cortical regionalization.
\end{compactenum}
The model's symbolic search generated computationally invalid and biologically implausible structures, demonstrating how insufficient domain context can lead to fundamental misalignment with both mathematical and biological principles underlying Boolean network dynamics.

\section{Analysis on Table~\ref{tab: full-symbolic}}
\label{appendix: corr_CDE}
Prior studies~\cite{shojaee2024llm, merler2024context} often use complexity as the sole standard for final selections of candidate predictions. However, we show that the generalization, represented by the performance during holdout evaluation (OOD), has a poor correlation with expression complexity. (a) As shown in Table~\ref{tab: CDE_correlations}, without introducing context, complexity may have a moderate correlation with $ACC_0.9$ when the dimension is small and not challenging. However, given samples with higher dimensions, the correlation became positive, meaning higher complexity can also give better generalization. (b) When context is introduced, besides the overall improved performance is observed as illustrated in \textbf{Obs. 4}, the correlation also turned positive from Dim=1 to Dim=3 and remained low for Dim=4. \textit{Such an obvious change potentially suggests that context is a more effective criterion for candidate ranking and selection during both iterative refinement and final evaluation.}

\begin{table}[htbp]
\small
\centering
\caption{Correlations between complexity and OOD $ACC_{0.9}$ across four dimensions with and without context}
\label{tab: CDE_correlations}
\begin{tabular}{l
                S[table-format=+1.3]
                S[table-format=+1.3]
                S[table-format=+1.3]
                S[table-format=+1.3]}
\toprule
\textbf{Condition} & \textbf{Dim 1} & \textbf{Dim 2} & \textbf{Dim 3} & \textbf{Dim 4} \\
\midrule
w/o context & -0.672 & -0.384 & -0.425 &  0.584 \\
w/  context &  0.097 &  0.728 &  0.165 & -0.174 \\
\bottomrule
\end{tabular}
\end{table}

\section{Hybrid Method}
\label{appendix: hybrid_method}

\begin{figure}[hbt]
    \centering
    \begin{subfigure}[b]{0.48\textwidth}
        \centering
        \includegraphics[width=0.7\textwidth]{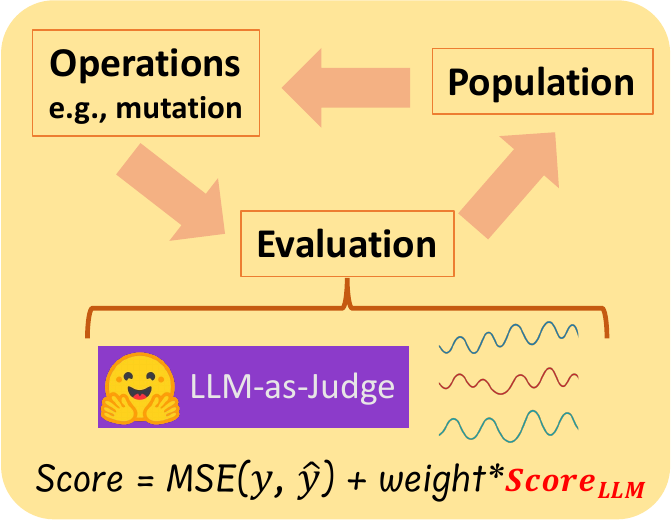}
        \caption{Hybrid method using Genetic Programming + LLM-as-Judge.}
        \label{fig: hybrid_gp}
    \end{subfigure}
    \hfill
    \begin{subfigure}[b]{0.48\textwidth}
        \centering
        \includegraphics[width=0.7\textwidth]{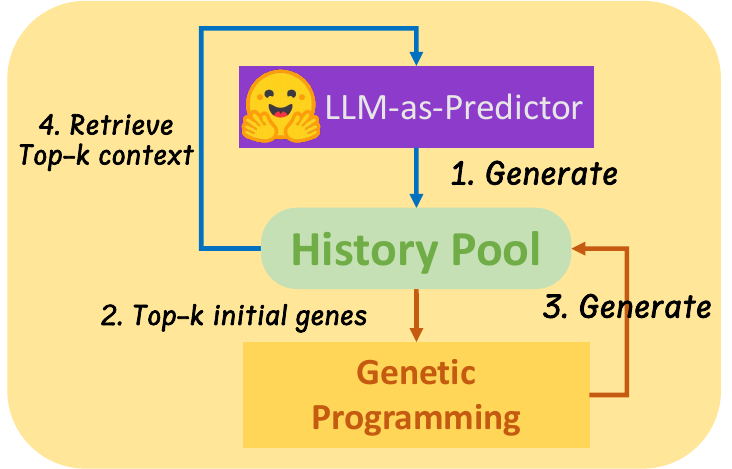}
        \caption{Hybrid method using Genetic Programming + LLM-as-Predictor, where GP helps expand the history pool with the current best expressions as initial population.}
        \label{fig: hybrid_llm}
    \end{subfigure}
    \caption{Hybrid method}
    \label{fig: hybrid_method}
\end{figure}

We adopt two frameworks for hybrid approach by separately let make GP and LLMs play different roles. (a) As shown in Figure~\ref{fig: hybrid_gp}, in addition to the quantitative evaluation using MSE, an additional qualitative score produced by LLM is incorporated in the evolution loop; (b) As shown in Figure~\ref{fig: hybrid_llm}, in addition to the original closed loop (\textcolor{blue}{blue line}), we provide an extended path (\textcolor{brown}{brown line}) that utilize GP to expand the history pool. From a different perspective, the initial population produced by LLMs also improves the generation for GP by providing context-enhanced initial populations.

\section{Convergence Rate}
\label{appendix: converge_rate}
\begin{figure}[hbt]
    \centering
    \begin{subfigure}[b]{0.4\textwidth}
        \includegraphics[width=\linewidth]{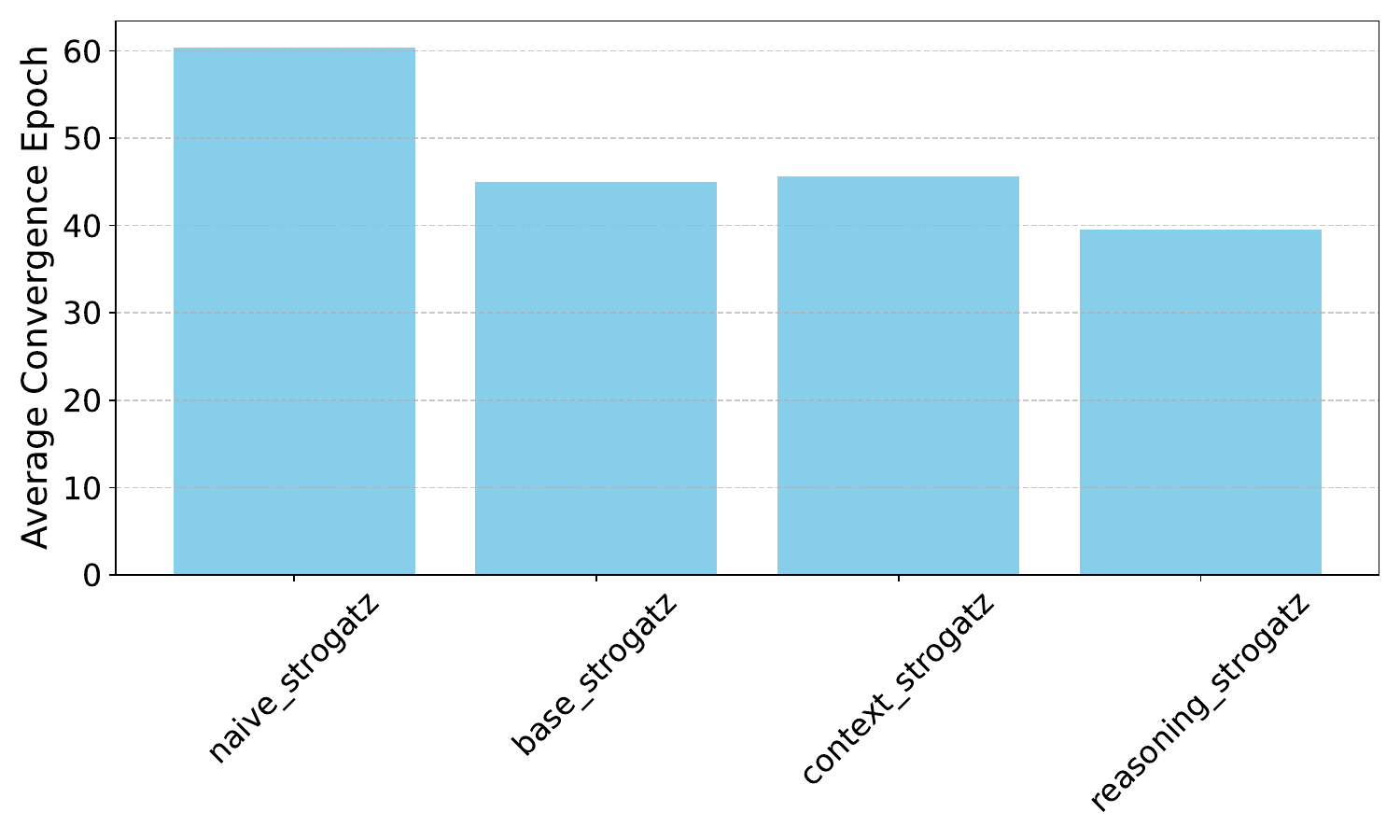}
        \caption{CDE}
    \end{subfigure}
    \caption{Convergence Rate on the CDE dataset.}
    \label{fig: convergence_rate}
\end{figure}

We further examine the convergence rate under various settings. As shown in Figure~\ref{fig: convergence_rate}, introducing both context and reasoning leads to a faster convergence rate on the CDE and BN datasets. In contrast, the SCM dataset exhibits an overall fast convergence rate across all settings, with context having only a marginal improvement.

\onecolumn
\section{Example Outputs}
\subsection{Example predictions across three tasks}
\label{appendix: pred_examples}

We present examples across three tasks in Table~\ref{tab: dataset_examples}.

\begin{figure*}[h]
    \centering
    \includegraphics[width=0.92\linewidth]{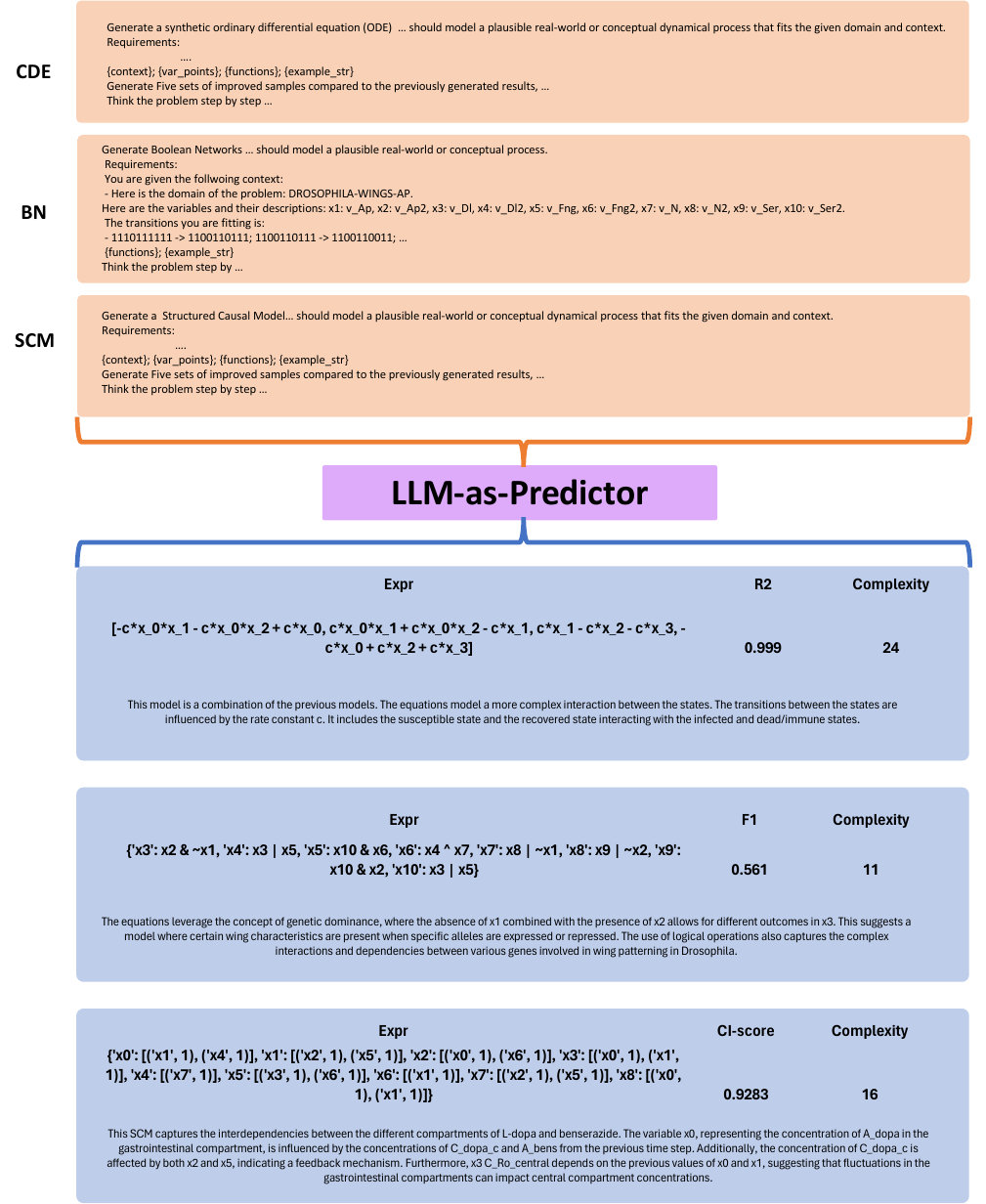}
    \caption{Example input and output across three tasks using LLMs-as-Predictors. The output is processed through verification.}
    \label{fig: tasks_examples}
\end{figure*}

\subsection{Candidates of CoT reasoning output on coupled-differential equations}
We show the list of candidates found for fitting the time series generated from the SEIR model in Table~\ref{tab: SEIR_candidates}.
\label{appendix: CoT_example}

\begin{sidewaystable}[h]
\scriptsize
\caption{Example candidates of a 4-dimensional coupled differential equation during inference.}
\label{tab: SEIR_candidates}
\setlength{\tabcolsep}{2pt}
\begin{tabularx}{\textheight}{@{}c|p{3cm}|c|c|p{5cm}|c|c|c|c@{}}
\textbf{\#} & \textbf{expr} & \textbf{R2} & \textbf{Complexity} & \textbf{reasoning (Qwen2.5-14B)} & \textbf{context\_alignment} & \textbf{scientific\_plausibility} & \textbf{conciseness\&clarity} & \textbf{logical\_coherence} \\
\hline
0 & \{-c*x\_0*x\_1 - c*x\_0*x\_2 + c*x\_0,\newline c*x\_0*x\_1 + c*x\_0*x\_2 - c*x\_1,\newline c*x\_1 - c*x\_2 - c*x\_3,\newline -c*x\_0 + c*x\_2 + c*x\_3\} & 0.99999 & 24 & This model combines previous models. It introduces\newline complex interaction between states. Transitions are\newline governed by rate constant c. Susceptible and\newline recovered states interact with infected/immune. & 4 & 4 & 3 & 4 \\
\hline
1 & \{c*x\_0*x\_1 - c*x\_2 + c*x\_3,\newline c*x\_1 + c*x\_2 - c*x\_3,\newline -c*x\_0 + c*x\_2 + c*x\_3,\newline c*x\_0 - c*x\_1 + c*x\_3\} & 0.99976 & 21 & x\_0 and x\_1 interact to affect x\_2. Other variable\newline pairs interact linearly. This might reflect a chain\newline of influence among factors—each influencing the\newline next in a progression. & 4 & 3 & 2 & 3 \\
\hline
2 & \{c*x\_0 + c*x\_1 - c*x\_2,\newline c*x\_1 + c*x\_2 - c*x\_3,\newline -c*x\_0 + c*x\_2 + c*x\_3,\newline c*x\_0 - c*x\_1 + c*x\_3\} & 0.99975 & 20 & This system maintains cyclic balance among variables.\newline Represents processes like ecological cycling where\newline each state influences and is influenced in turn,\newline sustaining continuous transitions. & 2 & 2 & 3 & 2 \\
\hline
3 & \{c*x\_0*x\_1 - c*x\_0*x\_2,\newline c*x\_0*x\_2 - c*x\_1,\newline c*x\_1 + c*x\_2 - c*x\_3,\newline -c*x\_0 + c*x\_3\} & 0.99825 & 17 & A variant epidemiological model. Introduces a slightly\newline more intricate structure for transitions. Dynamics\newline still guided by rate constant c. Greater attention\newline to susceptible and infectious interplay. & 4 & 4 & 3 & 4 \\
\hline
4 & \{c*x\_0*x\_1 - c*x\_2 + c*x\_3,\newline -c*x\_0*x\_2 + c*x\_1 + c*x\_3,\newline c*x\_0 - c*x\_1*x\_3 + c*x\_2,\newline -c*x\_0*x\_2 + c*x\_1 + c*x\_3\} & 0.99292 & 24 & All variables appear in every equation. Introduces\newline nonlinear terms, enabling feedback loops. May model\newline biochemical systems with mutual influence and\newline delayed or amplified effects. & 4 & 3 & 2 & 3 \\
\hline
5 & \{c*x\_0*x\_1 - c*x\_2,\newline c*x\_1*x\_2 - c*x\_3,\newline -c*x\_0*x\_1 + c*x\_2,\newline c*x\_0 - c*x\_1 + c*x\_3\} & 0.96474 & 17 & Models disease spread (SIR). x0: susceptible,\newline x1: exposed, x2: infected, x3: recovered. Each\newline equation captures standard transitions. E.g.,\newline x0*x1 for infections; x2 affects x3. & 5 & 4 & 4 & 4 \\
\hline
6 & \{c*x\_0 - c*x\_1*x\_2,\newline c*x\_1*x\_2 - c*x\_3,\newline -c*x\_0 + c*x\_2*x\_3,\newline -c*x\_0 + c*x\_1 + c*x\_3\} & 0.90896 & 17 & Similar to the SIR model with slight variations.\newline Includes multiplicative terms between states.\newline x0 and x1 drive infection; x2 and x3 influence\newline recovery rates. Captures more nuanced behavior. & 4 & 3 & 4 & 4 \\
\hline
7 & \{-c*x\_0*x\_1 - c*x\_0*x\_2 + c*x\_0,\newline c*x\_0*x\_1 - c*x\_1,\newline c*x\_0*x\_2 - c*x\_2,\newline c*x\_1 + c*x\_2 - c*x\_3\} & 0.79325 & 20 & Basic epidemiological structure (SIR). Variables:\newline x0: susceptible, x1: infected, x2: recovered,\newline x3: immune/dead. Rate constant c influences all\newline transitions between compartments. & 4 & 4 & 3 & 4 \\
\hline
8 & \{-c*x\_0*x\_1 + c*x\_0*x\_2 - c*x\_2,\newline c*x\_0*x\_1 - c*x\_1 - c*x\_2,\newline c*x\_2 - c*x\_3,\newline -c*x\_0*x\_2 + c*x\_1\} & 0.33899 & 20 & Represents SEIR model. dx0/dt: loss via infection,\newline dx1/dt: gain from infection, loss via exposure,\newline dx2/dt: gain from exposure, loss via recovery,\newline dx3/dt: gain from recovery, potential loss. & 4 & 4 & 3 & 4 \\
\end{tabularx}
\end{sidewaystable}

\subsection{RLM output for one sample with marked reasoning steps}
\label{appendix: RLM_example}

{\small
(\noindent\textbf{Step 1: Problem Setup})
Okay, let's see. I need to generate a system of coupled ODEs for the calcium\_dynamics domain. The variables are $x_0, x_1, x_2, x_3$, each with their descriptions as given. The time series data is provided, which includes $x_0$ values and the other variables. (\noindent\textbf{Step 2: Initial Observations}) First, the variables $x_0, x_1, x_2, x_3$ are each in the range of their description. The domain is calcium dynamics, so the equations need to model some process related to calcium concentration. Looking at the time series for $x_0$, which is a negative value, and the other variables being positive. Since calcium concentrations are usually positive, maybe the model is about the flow of calcium through the cell membrane, or the impact of different ions on calcium levels. The time series for $x_1$ is all zeros except at some points. The $x_1$ values are all 0.3, then increase. Maybe this is a transient phase or a step change. For $x_2$ and $x_3$, they seem to be varying linearly. (\noindent\textbf{Step 3: First Model Attempt}) Looking at the equations, perhaps there are some nonlinear terms. The variables can be combined using trigonometric functions. Let me think about a possible model. If I consider a calcium channel model, the rates of change could involve the concentration of ions. For example, the rate of change of $x_0$ might depend on $x_1, x_2,$ and $x_3$. Similarly, $x_1$ might relate to $x_2$ and $x_3$. (\noindent\textbf{Step 4: Linear Combination Trial}) But the problem is to create a model that fits the data. Let's think of a simple model. Suppose that $x_0 = c(x_1 + x_2 + x_3)$. Similarly, maybe $x_1 = c(x_2 - x_3)$. For $x_2$, maybe it's a linear function: $x_2 = c(x_0 + x_3)$. Similarly, $x_3 = c(x_3 - x_0)$. Putting it all together, the system would be:
\begin{verbatim}
c*x1 + c*x2 + c*x3 | c*x2 - c*x1 | c*x0 + c*x3 | c*x3 - c*x0
\end{verbatim}
(\noindent\textbf{Step 5: Plugging in Sample Data}) In the first sample: $x_0 = -43.0$, $x_1 = 0.03$, etc.

$x_0 = c(0.03 + 0.1 + 0.434) = c(0.564) \Rightarrow c = \frac{-43.0}{0.564} \approx -76.5$

$x_1 = c(0.1 - 0.434) = c(-0.334) \Rightarrow c = \frac{0.03}{-0.334} \approx -0.09$

Inconsistent values suggest equation mismatch. (\noindent\textbf{Step 6: Revising the Model}) Try again with the same structure: 

$c = \frac{x_0}{x_1 + x_2 + x_3} = \frac{-43.0}{0.564} \approx -76.5$  
$c = \frac{0.03}{0.1 - 0.434} = \frac{0.03}{-0.334} \approx -0.09$

Still inconsistent. (\noindent\textbf{Step 7: Trying Differential Form}) Suppose:
\begin{verbatim}
dx0/dt = c*(x1 + x2 + x3)
dx1/dt = c*(x2 - x3)
dx2/dt = c*(x0 + x3)
dx3/dt = c*(x3 - x0)
\end{verbatim}

(\noindent\textbf{Step 8: Numerical Check}) With $x_0 = -43.0, x_1 = 0.03, x_2 = 0.1, x_3 = 0.434$:

\begin{verbatim}
dx0/dt = c*0.564
dx1/dt = c*(-0.334)
dx2/dt = c*0.464
dx3/dt = c*43.434
\end{verbatim}

Still inconsistent c values. (\noindent\textbf{Step 9: Considering Other Forms}) Maybe use other terms, e.g., absolute values or trigonometric forms.

\noindent\textbf{Step 10: Trying Absolute Values}  
\begin{verbatim}
dx0/dt = |x1 + x2 + x3|
dx1/dt = |x2 - x3|
dx2/dt = |x1 - x3|
dx3/dt = |x3 - x0|
\end{verbatim}

Still not matching data. (\noindent\textbf{Step 11: Simpler Model Revisited}) Try:

\begin{verbatim}
dx0/dt = c*x1 + c*x2 + c*x3
dx1/dt = c*x2 - c*x3
dx2/dt = c*x0 + c*x3
dx3/dt = c*x3 - c*x0
\end{verbatim}

(\noindent\textbf{Step 12: Integrating into Sample}) With $c = 1$:

\begin{verbatim}
dx0/dt = 0.564
dx1/dt = -0.334
dx2/dt = -42.566
dx3/dt = 43.434
\end{verbatim}

(\noindent\textbf{Step 13: Final Justification}) Suggests calcium dynamics via flow of ions. $x_0$ as calcium concentration, $x_1, x_2, x_3$ as inflow/outflow.

(\noindent\textbf{Step 14: Final Model Form and JSON}) 
\begin{verbatim}
dx0/dt = c*x1 + c*x2 + c*x3
dx1/dt = c*x2 - c*x3
dx2/dt = c*x0 + c*x3
dx3/dt = c*x3 - c*x0
\end{verbatim}

(\noindent\textbf{Step 15: Format Verification}) Follow pipe-separated format:

\begin{verbatim}
{"eq": "c*x_0*x_1 + c*x_2 - c*x_3 | c*x_1/x_0 + c*x_3 | c*x_2 - c*x_0*x_3 | c*x_3 + c*x_0 - c*x_1",  "dim": 4}
\end{verbatim}

(\noindent\textbf{Final JSON Output})

\begin{verbatim}
{
  "eq": "c*x_0*x_1 + c*x_2 - c*x_3 | c*x_1/x_0 + c*x_3 | c*x_2 - c*x_0*x_3 | c*x_3 + c*x_0 - c*x_1",
  "dim": 4
}
\end{verbatim}
}

\clearpage
\section{Prompts}
\label{appendix: prompts}

\begin{figure}[hbt]
    \centering
    \includegraphics[width=\linewidth]{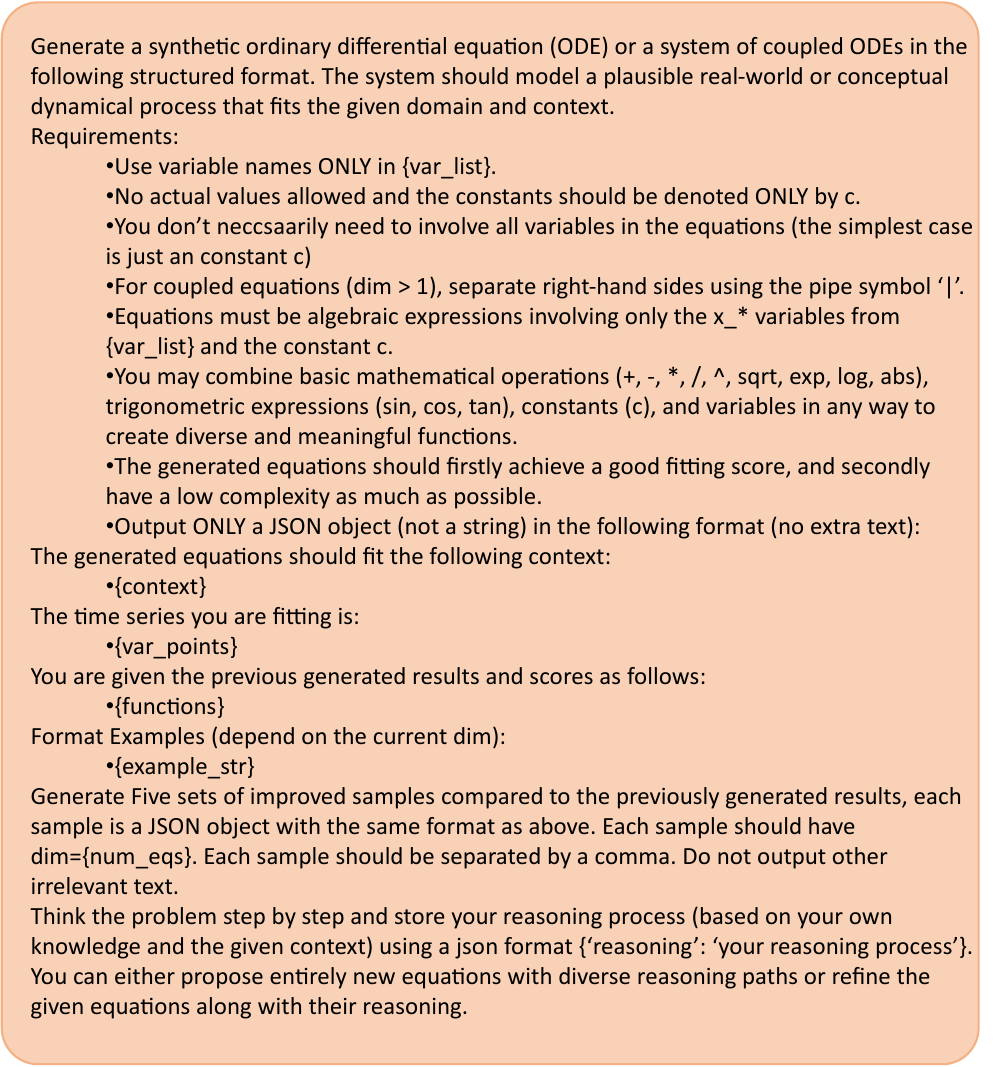}
    \caption{Prompt with CoT and Context for CDEs.}
    \label{fig: DE_prompt}
\end{figure}

\begin{figure}[hbt]
    \centering
    \includegraphics[width=\linewidth]{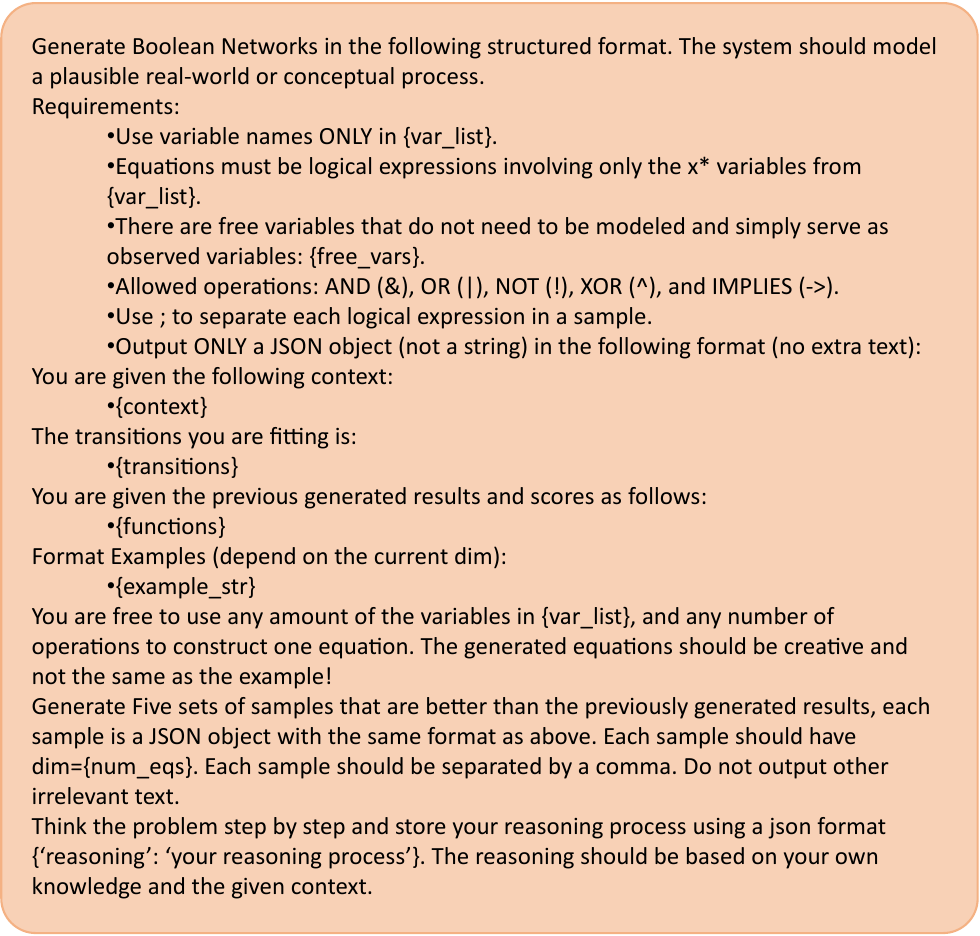}
    \caption{Prompt with CoT and Context for BNs.}
    \label{fig: BN_prompt}
\end{figure}

\begin{figure}[hbt]
    \centering
    \includegraphics[width=\linewidth]{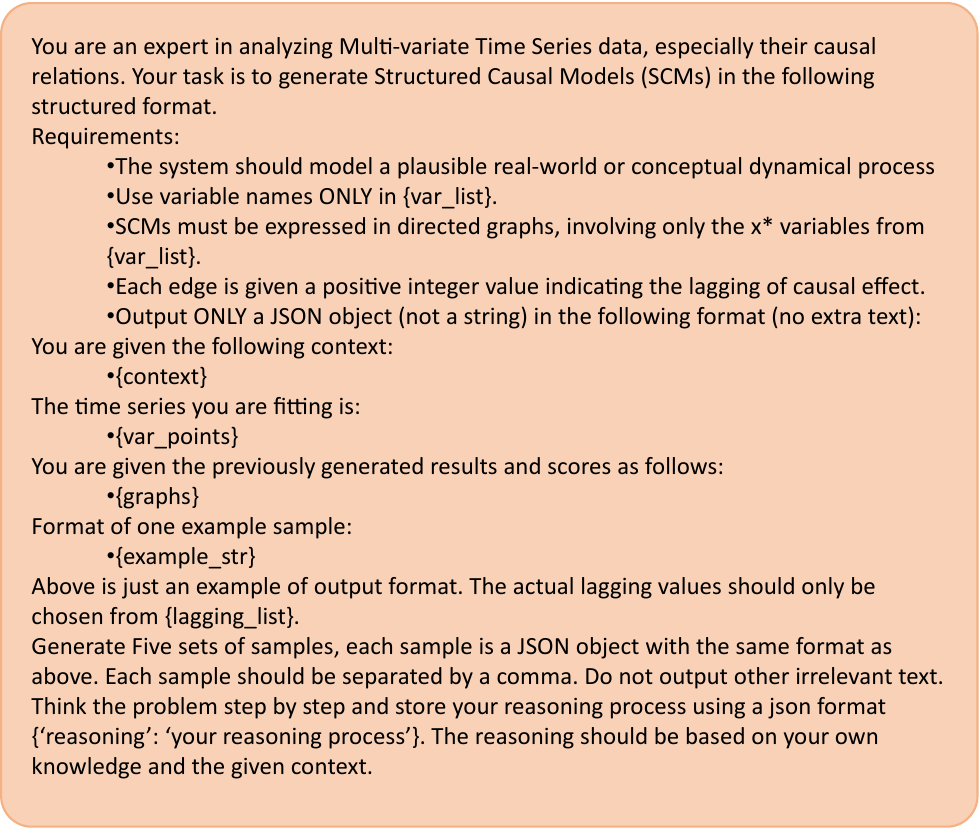}
    \caption{Prompt with CoT and Context for SCMs.}
    \label{fig: SCM_prompt}
\end{figure}

\begin{figure}[hbt]
    \centering
    \includegraphics[width=\linewidth]{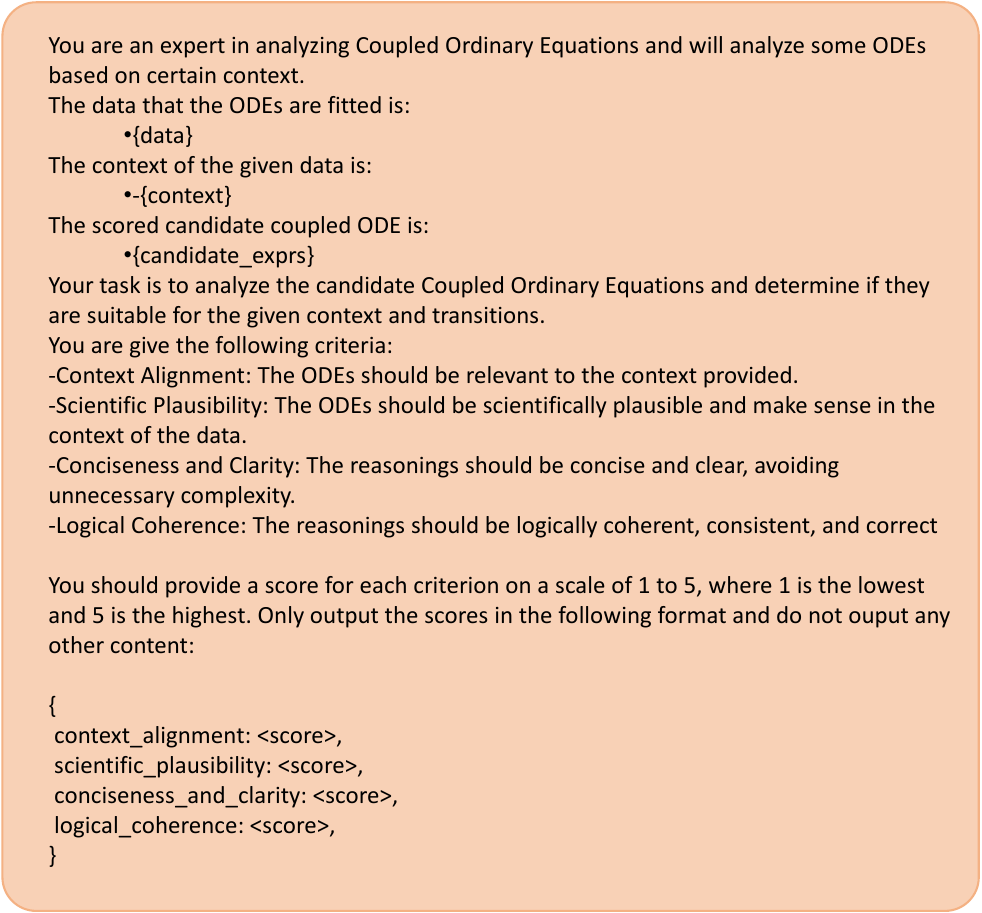}
    \caption{LLM-as-Judge prompt for CDEs.}
    \label{fig: DE_judge_prompt}
\end{figure}

\end{appendices}

\end{document}